\documentclass[letterpaper, 10 pt, conference]{ieeeconf}

\IEEEoverridecommandlockouts 
\usepackage{dblfloatfix}
\usepackage[noadjust]{cite}
\usepackage{amsmath,amssymb,amsfonts}
\usepackage{algorithmic}
\usepackage{algorithm}
\usepackage{fancyhdr}
\usepackage{graphicx}
\usepackage{textcomp}
\usepackage{xcolor}
\usepackage{pifont}
\usepackage{multicol}
\usepackage{verbatim}
\usepackage{pdflscape}
\usepackage{booktabs}
\usepackage{wrapfig}
\usepackage{multirow}
\usepackage[citecolor=black,colorlinks=true]{hyperref}
\usepackage[caption=false,font=footnotesize]{subfig}
\usepackage[normalem]{ulem}

\newcommand\edit[1]{\textcolor{black}{#1}}

\begin{document}

\title{Inverse k-visibility for RSSI-based Indoor Geometric Mapping}

\author{Junseo Kim\textsuperscript{*}, Matthew Lisondra\textsuperscript{*}, Yeganeh Bahoo, and Sajad Saeedi, 
\thanks{\textsuperscript{*} These authors contributed equally to this work.}
\thanks{Junseo Kim is with the Department of Cognitive Robotics, Delft University of Technology, Delft, The Netherlands (e-mail: j.kim-18@student.tudelft.nl)}
\thanks{Matthew Lisondra is with the Department of Mechanical and Industrial Engineering, University of Toronto, Toronto, ON (e-mail: lisondra@mie.utoronto.ca)}
\thanks{Yeganeh Bahoo is with 
the Department of Computer Science, Toronto Metropolitan University, Toronto, ON (e-mail: bahoo@torontomu.ca).}
\thanks{Sajad Saeedi is with the Department of Computer Science, University College London, London, UK (e-mail: s.saeedi@ucl.ac.uk)}}

\maketitle

\fancypagestyle{firstpage}{
    \fancyhf{} 
    \fancyhead[L]{} 
    \renewcommand{\headrulewidth}{0pt}
}
\thispagestyle{firstpage}


\begin{abstract}
    
In recent years, the increased availability of WiFi in indoor environments has gained interest in the robotics community to utilize WiFi signals for indoor simultaneous localization and mapping algorithms. 
This paper discusses the challenges of achieving high-accuracy geometric map building using WiFi signals. The paper introduces the concept of \emph{inverse k-visibility}, developed from the k-visibility algorithm, to identify free space in an unknown environment, used for planning, navigation, and obstacle avoidance.  Comprehensive experiments, including those utilizing single and multiple RSSI signals, were conducted in both simulated and real-world environments to demonstrate the robustness of the proposed algorithm. Additionally, a detailed analysis comparing the resulting maps with ground-truth LiDAR-based maps is provided to highlight the algorithm's accuracy and reliability.
\end{abstract}

\section{Introduction}
\label{sec:introduction}
Simultaneous localization and mapping (SLAM) is an essential process, involving self-localizing and the construction of a map of the environment, especially important for autonomous robots.
%
Current state-of-the-art approaches rely heavily on sensors like cameras, inertial measurement units (IMUs), laser range finders, and ultrasound. These sensors provide data that is used to generate accurate estimates of the environment map. 
However, 
cameras, laser range finders, and ultrasound sensors can be obstructed, corrupted by noise, or cause privacy concerns. 
Therefore, alternative approaches are essential when these exteroceptive sensors are unreliable or insufficient. Developing robust methods that can compensate for these limitations is critical to improving overall performance and ensuring accurate environment mapping in diverse and challenging conditions.

\begin{figure}
  \centering
  \includegraphics[width=0.75\linewidth]{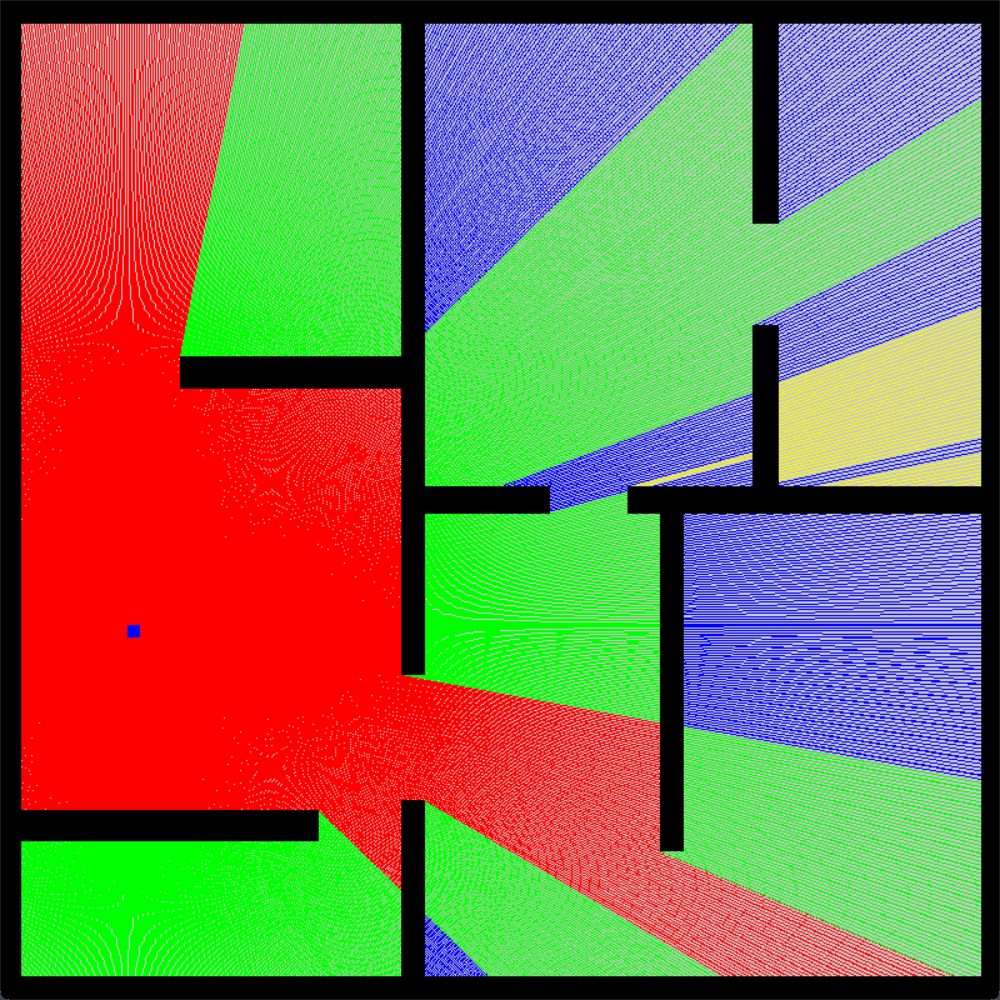}
  \caption{{The concept of $k$-visibility is demonstrated where different $k$-values are shown: $k=0$ (red), $k=1$ (green), $k=2$ (black) and $k=3$ (yellow). Based on a straight-line measure from a reference point such as a router (shown in dark black) to a desired location, $k$-visibility is a metric on how many times the line traverses through a wall/obstacle.}}
  \label{fig:k_visibility}
\end{figure}

As WiFi networks occupy many indoor and public spaces, the use of information extracted from WiFi signals is a promising alternative for SLAM. One such information is known as the received signal strength indicator (RSSI), indicating the strength of the signal received by a WiFi receiver. RSSI is a function of the geometry of the environment and the location of the WiFi emitter and receiver. 
WiFi is particularly advantageous in situations where traditional sensors such as cameras, lasers, and ultrasound may be inadequate, such as in environments with privacy concerns or poor lighting conditions~\cite{kudo2017utilizing}. Methods such as WiFiSLAM~\cite{ferris2007wifi} have predominantly concentrated on WiFi-based localization and the potential estimation of WiFi router positions, without generating a geometric map of the environment, needed for many robotic applications. 
There have been works relying on crowdsourcing~\cite{zhou2021crowdsourcing}, 
{showing that it is possible to generate an environment map using WiFi data. However, these methods often face challenges such as limited accuracy due to signal interference, variability in signal strength, and the need for extensive data from multiple users for reliable results. }
Additionally, processing of the map with primarily RSSI information in the algorithms of many works shows that it fluctuates unstably, being unreliable, particularly in environments that are dynamically changing~\cite{arun2022p2slam}.
{
Thus, a gap exists in developing a reliable WiFi-based mapping technique. 
The main problem arises in acquiring, from the WiFi data, an accurate geometric shape of the environment.}
This paper\footnote{This paper is partly based on our earlier paper~\cite{kim2024structure}, extended with multiple routers, metric evaluations, and more experiments.} presents a novel technique capable of generating geometric maps from WiFi signals. The method utilizes developments in computational geometry, namely the {\it $k$-visibility} algorithm~\cite{o1987art}, see Fig.~\ref{fig:k_visibility}. We extend the $k$-visibility by presenting the concept of {\it inverse $k$-visibility}. This novel concept is then utilized to develop a geometric map of an unknown environment from WiFi signals, coined Structure from WiFi (SfW).



Knowing accurately the mapped free space is imperative for the control of many autonomous systems, which need to know how to plan paths, making sure not to collide with occupied and obstacle spaces or to minimize the number of collisions. The contributions of our work are:\footnote{\href{https://sites.google.com/view/structure-from-wifi/journal-extension}{https://sites.google.com/view/structure-from-wifi/journal-extension}}
1)~A novel algorithm that is capable of generating geometric maps using WiFi signals received from multiple routers, 2)~benchmarking the WiFi-generated maps with LiDAR-generated maps by comparing the area, number of data points, $k$-value prediction True/False setting, $k$-value accuracy percentage, IOU and MSE scores, and 3)~evaluation on real-world collected from indoor spaces.


The rest of this work is organized as follows: 
Sec.~\ref{sec:related} presents an overview of recent approaches to WiFi-based SLAM and challenges faced in applications. 
Sec.~\ref{sec:background} describes the background of the $k$-visibility algorithm.
Sec.~\ref{sec:dense} proposes the inverse $k$-visibility algorithm.
%
Sec.~\ref{sec:sparse} extends the inverse $k$-visibility algorithm to real-world experiments utilizing sparse information.
Sec.~\ref{sec:exp} presents the experimental results. 
Sec.~\ref{sec:dis} discusses current challenges and limitations of the work.
\color{black}
Finally, Sec.~\ref{sec:con} discusses future works and concludes the paper.

\section{Literature Review}\label{sec:related}
The following section reviews recent progress in WiFi-only SLAM algorithm and wireless-based mapping technologies, pointing out the challenges encountered in generating a precise environment map from signal-strength data.

  
WiFi-based localization techniques have seen extensive adoption in recent years. Ferris {\it et al.}~\cite{ferris2007wifi} succeeded in localization by converting high-dimensional signal strength data into a two-dimensional latent space and resolving SLAM using Gaussian Process Latent Variable Modeling (GP-LVM). This work was followed by other studies, including, \cite{xiong2017diversified, miyagusuku2016improving}. Graph-based algorithms for SLAM have also been explored, as in ~\cite{huang2011efficient, liu2019collaborative, herranz2016wifi, arun2022p2slam}. Enhancements in WiFi observation models have shown potential for high-accuracy mapping, as described in the research by \cite{kudo2017utilizing, he2015wi}.
Both learning and non-learning methods have their advantages and disadvantages, depending on the robotic system.
We will first address learning-based WiFi map construction approaches (See Sec.~\ref{sec:learningapproaches}), and go through non-learning, geometric WiFi-map methods (See Sec.~\ref{sec:nonlearning})

\subsection{Learning-Based WiFi Methods}\label{sec:learningapproaches}

The prominence of learning-based WiFi map construction has alleviated many issues with conventional geometric methods, but with the disadvantage of increased computational need in its pre- and post-processing steps ~\cite{Bellavista1}, \cite{LiuFen1}. Zou et al. ~\cite{zou1} introduced WiGAN, a generative adversarial network (GPR-GAN) for constructing detailed indoor radio maps using Gaussian process regression (GPR) with received signal strengths (RSS). Other notable works include ~\cite{Ayyalasomayajula1}, ~\cite{Xuhen1}, ~\cite{Karmanov1}, and ~\cite{chenXi1}. Zhang et al. ~\cite{ZhangLe1} proposed a deep fuzzy forest to address the limitations of decision trees and deep neural networks in end-to-end training for WiFi-based indoor robot positioning. Additionally, Ayinla et al. ~\cite{Ayinla1} developed SALLoc, a WiFi fingerprinting indoor localization scheme using Stacked Autoencoder (SAE) and Attention-based Long Short-Term Memory (ALSTM), which demonstrated high accuracy and robustness in large-scale indoor environments. Wang et al. ~\cite{Wang1} introduced MapLoc, an LSTM-based indoor localization system that uses uncertainty maps created from magnetic field readings and WiFi RSS, showing superior performance in location prediction. Park et al. ~\cite{Park1} reviewed various machine learning techniques for WiFi-based indoor positioning, emphasizing their potential to enhance accuracy and scalability. Turgut and Gorgulu Kakisim ~\cite{Turgut1} proposed a hybrid deep learning architecture combining LSTM and CNN for WiFi-based indoor localization, achieving higher accuracy than baseline methods.

\subsection{Non-Learning WiFi Methods}\label{sec:nonlearning}

The alternative to deep learning (DL) and machine learning (ML) modelling deals with non-learning methods. 
Several studies have used the sensors integrated into smartphones to infer indoor floor plans through a crowdsensing approach~\cite{zhou2021crowdsourcing}. Research in this area often utilizes inertial sensors alongside WiFi signal strength data, both of which are common in commercial smartphones. Key papers include ~\cite{luo2014piloc, shin2011unsupervised, zhou2015alimc, zhou2018graph, shen2013walkie, jiang2013hallway, alzantot2012crowdinside, liang2016sensewit}. The main limitation of these studies is that they only produce ``traversable maps", identifying navigable areas while leaving occupied zones uncharted. Additionally, these works rely on crowdsourced or public data, which may not always be accessible, thereby restricting their applicability.  

In a different approach, Gonzalez-Ruiz and Mostofi~\cite{gonzalez2013cooperative} introduced a method to generate a non-invasive occupancy grid map using wireless measurements and directional antennas. Their framework employed a coordinated robot setup to create a 2D map of an environment by collecting data through walls and other obstacles. Similar techniques have been applied in Ultra Wideband (UWB) SLAM studies, utilizing UWB signal path propagation models with directional antennas to achieve similar goals~\cite{deissler2010uwb, deissler2012infrastructureless}. 

Further, non-learning methods that do not rely on model predictions, such as Tong et al. ~\cite{tong1}, achieve decimeter-level localization accuracy by segmenting Wi-Fi access points (APs) into groups. These methods generate local maps from the APs and then merge them to create a comprehensive global Wi-Fi map. Ninh et al. ~\cite{Ninh1} proposed a random statistical method for WiFi fingerprinting, achieving a maximum positioning error of less than 0.75 meters. Shi et al. ~\cite{Shi1} introduced a training-free indoor localization approach using WiFi round-trip phase and factor graph optimization, achieving a mean absolute error (MAE) of 0.26 meters. Estrada et al. ~\cite{Estrada1} developed an indoor positioning system using OpenWRT, achieving an average margin error of 2.43 meters. Tao and Zhao ~\cite{Tao1} presented algorithms utilizing extreme values for AP selection and positioning, demonstrating improved accuracy. Sulaiman et al. ~\cite{Sulaiman1} explored radio map generation methods to enhance RSSI-based indoor positioning, achieving approximately 0.45 meters accuracy. Azaddel et al. ~\cite{Azaddel1} proposed SPOTTER, a WiFi and BLE fusion method based on particle filtering, showing a 35\% improvement in accuracy. Tao et al. ~\cite{TaoYe1} introduced CBWF, a lightweight WiFi fingerprinting system achieving an average localization accuracy of 2.95 meters. Tang et al. ~\cite{Tang1} combined WiFi and vision for map construction and maintenance, enhancing accuracy and stability. Finally, Jurdi et al. ~\cite{Jurdi1} presented WhereArtThou, a WiFi-RTT-based indoor positioning system, achieving 90th percentile distance errors of 1.65 meters, suitable for commercial deployment.

\color{black}

\section{Background: \texorpdfstring{$k$-visibility}{k-visibility}}\label{sec:background}

The concept of {\it $k$-visibility} offers a unique perspective on WiFi systems and will be used throughout our algorithms. This concept was initially introduced as the {\it modem illumination problem}~\cite{fabila2009modem,aichholzermodem}. $k$-visibility investigates the extent of the area that can be observed from the router points at the vertices of a polygon. This concept is an extension of the basic \emph{visibility} problem, which focuses on determining the region that a single point within a polygon can see~\cite{o1987art}. 
In a simple polygon \emph{P}, two points \emph{p} and \emph{q} are considered to be mutually visible if the straight line connecting them does not intersect the exterior of \emph{P}~\cite{o1987art}. The visibility idea came from the Art Gallery Problem posed by Victor Klee in 1973~\cite{klee1969every}, which seeks to determine the minimum number of guards required to monitor an art gallery with \emph{n} walls. The area that can be seen from a specific point forms what is known as a \emph{visibility polygon}~\cite{o1987art}.



On the other hand, two points \emph{p} and \emph{q} are defined as \emph{$k$-visible} if the line segment connecting them intersects the exterior of polygon \emph{P} no more than \emph{k} times~\cite{o1987art}. Consequently, for a given \emph{k}, it is possible to determine the $k$-visible region from a particular vantage point \emph{p}.

%
%
%

Fig.~\ref{fig:k_visibility} shows a map displaying various $k$-visibility values from the viewpoint of a point marked in dark black within the red area. Here, red cells indicate 0-visibility, green cells represent 1-visibility, black cells correspond to 2-visibility, and yellow cells denote 3-visibility.

%
The vantage point \emph{p} is referred to as a \emph{k-transmitter}, and $k$-visible polygons can be identified for each value in the set $\{0, 1, .., k\}$~\cite{o1987art}. Several algorithms exist to generate $k$-visibility plots, each with varying levels of computational complexity~\cite{bajuelos2012hybrid, bahoo2020computing, bahoo2019time}.

 \begin{figure}
  \centering
  \includegraphics[width=.9\linewidth]{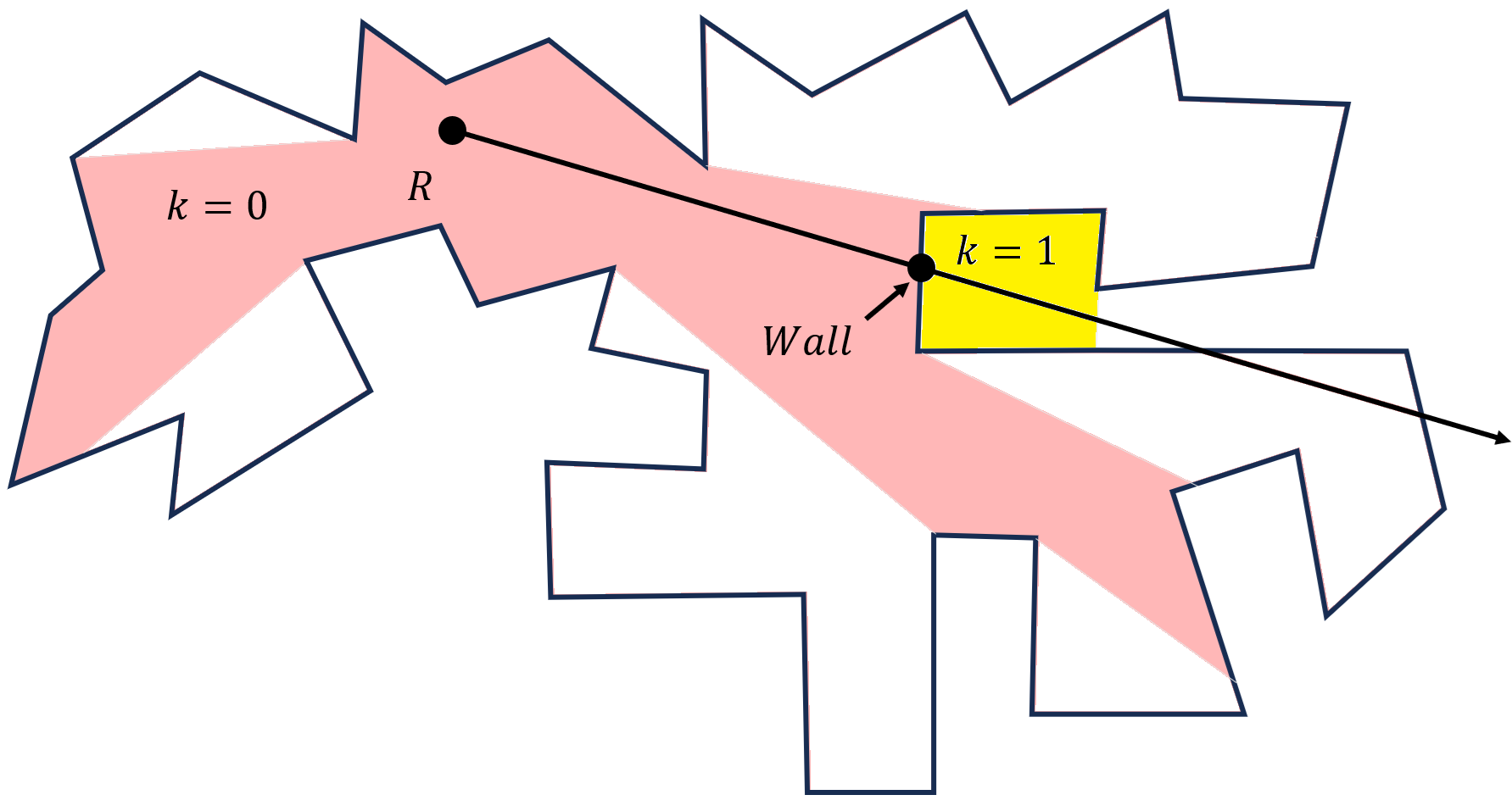}
  \caption{Diagram illustrating the ray-drawing principle fundamental to the inverse $k$-visibility algorithm.
  The wall is located where two consecutive $k$-value regions intersect along a ray projected from the router point. Only the region within the $k=1$ area is displayed.}  
  \label{fig:raydrawingprinciple}
\end{figure}

\begin{algorithm}[th]
\caption{Dense Inverse k-Visibility Algorithm}
\label{alg:inverse_k_visibility}
\begin{algorithmic}[1]
\STATE \textbf{Inputs:} 
\STATE \quad Reference point \( R \) (e.g., WiFi router position) \label{ln:input1}
\STATE \quad $k$-value of each point along the rays \label{ln:input2}
\STATE \textbf{Outputs:}
\STATE \quad A map \( M \) that classifies regions as free space or walls \label{ln:output}
\STATE \textbf{Initialize:}
\STATE Set the map \( M \) to unknown values (\( M_{ij} \gets 127 \) $\forall$ \( i,j \))\label{ln:ini-map}
\FORALL{rays \( r \in \mathcal{R} \) cast from the reference point \( R \)}\label{ln:for-each-ray}
    \FORALL{points \( p \) along ray \( r \)}\label{ln:for-each-p}
        \STATE get $k\text{-val}_p$ \label{ln:get-k-val}
        \IF{$k\text{-val}_p = 0$}\label{ln:k-is-zero}
            \STATE Mark the region between \( R \) and \( p \) as free in \( M \) \label{ln:markfree}
        \ELSIF{$k\text{-val}_p \geq 1$} \label{ln:k-le-one}
            \STATE Mark the point where \( k \) changes as a wall in \( M \)\label{ln:markwall}
        \ENDIF
    \ENDFOR
\ENDFOR
\STATE Return the final map \( M \)
\end{algorithmic}
\end{algorithm}

\section{\texorpdfstring{Dense Inverse $k$-visibility}{Dense Inverse k-visibility}} \label{sec:dense}

In this section, we will go over the preliminaries and concept of the proposed \emph{dense inverse $k$-visibility}, starting with an overview of the algorithm involved.

The $k$-visibility problem, explained in the previous sections, takes the location of a router and the geometry of the environment, e.g., the walls, as inputs. It then generates a value for each point in the space, which indicates how many walls exist between the point of the router. This value, known as the $k$-value, is calculated from RSSI in this paper. The inverse of this problem is of interest, where we assume that for each point in the space, we know its $k$-value as well as the location of the router. The objective is to determine the geometry of the environment, e.g., the walls. This is useful for making a geometric map of the environment for robotic applications. Since we assume the $k$-value is known for `all' points, the inverse problem is coined the `dense' inverse $k$-visibility. Mathematically, this can be written as:
\begin{itemize}
  \item $k$-visibility
  \begin{equation}
    \textcolor{black}{\text{k\text{-}val}_p = \text{k\_vis}(R, p \mid map),\ \forall\,p \in S},
  \end{equation}
  \item dense inverse $k$-visibility
  \begin{equation}
    \textcolor{black}{map(p) = \text{dense\_inv\_k\_vis}(R, \text{kval}_p, p),\ \forall\,p \in S},
  \end{equation}
\end{itemize}
\noindent where $S$ is the desired space (which is 2D in this paper), and $k\text{-}val_p$ is the $k$-value for point $p \in S$. $R$ is the 2D location of the router, and $map$ is the geometry (2D map) of the environment, represented as an occupancy grid map. \textcolor{black}{We write $\text{k\_vis}(R, p \mid map)$ to indicate evaluation with respect to the geometry.}

\color{black}If the entire set of points representing the change in $k$-values is found, the geometry of the environment map can be extracted by the inverse $k$-visibility algorithm. 
As shown in Fig.~\ref{fig:k_visibility}, the wall and occupied cells lie on the intersection between consecutive $(k_{i-1},k_i)$ portions of the $k$-visible map. 
{The $k$-visibility maps must be aligned first such that the location of the router is the same in all of them.} 


{Alg.~\ref{alg:inverse_k_visibility} shows the details of the `dense inverse $k$-visibility' algorithm. 
It works by reversing the traditional $k$-visibility approach, where $k$-value regions are used to generate a map of the environment. 
The inputs are the 2D position of the router and $k$-values of each point along the ray coming out of the router (lines \ref{ln:input1}, \ref{ln:input2}).
The output is a 2D occupancy map showing the free and occupied cells (line \ref{ln:output}). Every cell of the map is initialized as unknown (line \ref{ln:ini-map}). 
The algorithm begins by casting rays $\mathcal{R}$ from a reference point $R$ (line \ref{ln:for-each-ray}), and for each point $p$ along the ray (line \ref{ln:for-each-p}), the $k$-value is retrieved. The values are calculated from RSSI values, which will be explained in the next section. If the $k$-value at a point $p$ is zero (line \ref{ln:k-is-zero}), meaning no obstacles are encountered, the algorithm marks the entire region between $R$ and $p$ as free space or 255 (line \ref{ln:markfree}). On the other hand, if $k(p) \geq 1$ (line \ref{ln:k-le-one}), the algorithm identifies the point where the $k$-value changes and marks this as the location of an obstacle or 0 (line \ref{ln:markwall}). The algorithm iterates through all rays $r \in \mathcal{R}$ (line \ref{ln:for-each-ray}), progressively updating the map $M$. The final map $M$ is returned once all rays have been processed, providing a detailed map of the environment. The function that maps signal strength to $k$-values will be discussed in detail in Sec.~\ref{sec:sparse}.}

\begin{figure}[t]
  \centering
  \includegraphics[width=.8\linewidth]{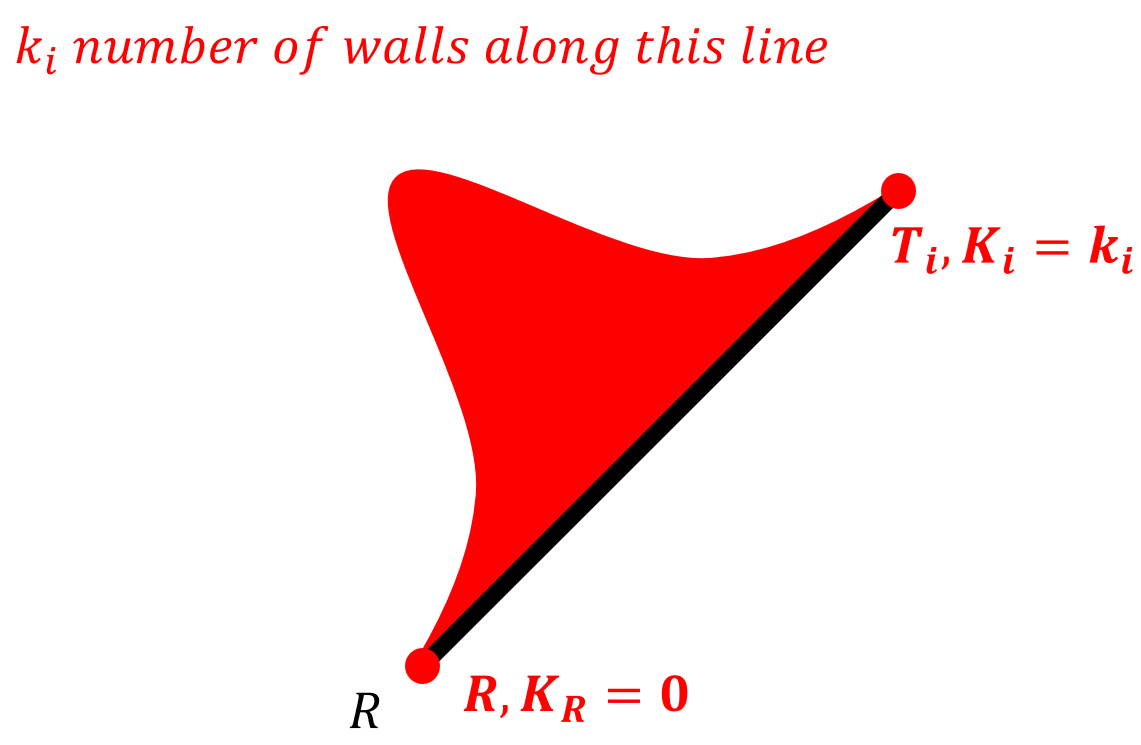}
  \caption{Ray-drawing for an arbitrary trajectory coordinates $T_i$ with an associated $k$-value $k_i$. According to the definition of $k$-visibility, the ray $\overline{RT_i}$ must intersect exactly $k_i$ walls along its path.}
  \label{fig:rti_line_drawing}
\end{figure} 

\color{black}
\section{\texorpdfstring{Sparse Inverse $k$-visibility}{Sparse Inverse k-visibility}}\label{sec:sparse}

The dense inverse k-visibility algorithm described in the previous section has a major limitation: the k-visibility maps need to be dense. In other words, extracting the $k$-value for the entire environment plot can often be too computationally demanding and renders the full approach not feasible, particularly concerning real-world applications. 
An alternative is to only do a subpart of the map, extracting instead \emph{sparse} $k$-values, this approach wherein here we call {\it sparse inverse $k$-visibility} is based on relating coordinates of known consecutive $k$-value coordinates located along the same casted ray from the router point to the trajectory. An estimate can be made of the router position and trajectory by dead reckoning using inertial measurement units or wheel encoders, whichever of the two is available for use in odometry.


We now go over the algorithm in greater detail. First, assume that we are in an indoor environment whose map we are trying to determine, assume that the router position is defined and well known, $R= (r_x, r_y)$. Assume that a user or mobile robot, which of whom is connected to this router signal's WiFi network, is traversing the indoor space with trajectory $T = [T_1, T_2, ..., T_{max}]$ where $T_i$ represents the $(x,y)$ position of a point along the $i^{th}$ coordinate in the trajectory. Assume that along with trajectory $T$, there is set $K=[K_1, K_2, ..., K_{max}]$ where $K_i$ represents the associated $k$-value from router position $R$ to trajectory in $i$-th step $T_i$, where $k$ is represented here as the integer representing the number of obstacles present in the casted ray from router $R$ to trajectory in this $i$-th step $T_i$.

This approach is probabilistic, and so \emph{probabilistic sparse inverse $k$-visibility} algorithm then works in three parts:


\begin{enumerate}
    \item Extracting $k$-values: for each coordinate, we have an associated $k$-value with respect to a known router position (See Sec.~\ref{subsec:k});
    \item Mapping Free Space: knowing when a space is obstacle or obstruction-free is defined (See Sec.~\ref{subsec:free});
    \item Mapping Occupied Space: the probabilistic approach is in classifying the obstacle or obstruction space, using a three-step process which is firstly to extract the casted ray, then do ray segmentation, apply Gaussian probability defining (See Sec.~\ref{subsec:occupied}).
\end{enumerate}


This probabilistic sparse inverse $k$-visibility algorithm assumes that the environment map is organized in a 2D grid format. This takes the form of an occupancy grid map style of mapping.
From ~\cite{thrun2002robotic}, occupancy grid mapping assumes perfectly cut tiles as subparts of the map environment. To make note of, under this assumption, the map is approximated and not continuous as it truly is, apart from a grid.
This then also would mean we make the assumption too that our casted ray array $\overline{RT_i}$'s elements are elements from tiles of a grid from router to trajectory. And based on $k$-visibility, this ray is bound to have a $k_i$ number of walls defined in its array. And so, if a space is deemed occupied, walls are then approximated as taking discrete filled-out tiles from this grid map environment. This concept is shown in Fig~\ref{fig:rti_line_drawing}.

%
%

\subsection{\texorpdfstring{Extracting $k$-values}{Extracting k-values}}\label{subsec:k}

\color{black}A wall prediction model was proposed by Fafoutis {\it et al.}~\cite{fafoutis2015rssi}, in which $k$-values were set based on RSSI measurements in a real experimental scenario, based on the trajectory traversed by a wearable sensor and a known access point. In this regard, the model is known to be an RSSI-based wall predictor whose function uses wearable sensor position as input to output the number of walls from this point to the known access point. This predicted number of walls has an upper bound of $K$ based on a sequence of RSSI bounds, $t_{1}, t_{2}, ..., t_{K}$ with function form ~\cite{fafoutis2015rssi}:
\begin{equation}
\label{eq:rssi_function}
f(P_{RSSI}) = 
\begin{cases} 
      0 & \text{if } P_{RSSI} > t_{1} \\
      1 & \text{if } t_{1} \geq P_{RSSI} > t_{2} \\
      \vdots & \\
      K-1 & \text{if } t_{K-1} \geq P_{RSSI} > t_{K} \\
      K & \text{if } t_{K} \geq P_{RSSI}
\end{cases}
\end{equation}

To note, RSSI signal strength weakens when passing through an obstacle, obstruction, or wall, and so the following is assumed:

\begin{equation}
    t_{K} < t_{K-1}: k\in [1, K]\label{eq:rssi_weakening_in_time}
\end{equation} 

We use the $K$-Means algorithm for the sake of brevity and simplicity, as the task is based on unsupervised learning and requires no training utilizing labelled data. This is a good approach for mapping from WiFi signals.


\begin{figure}[b]
  \centering
  \includegraphics[width=\linewidth]{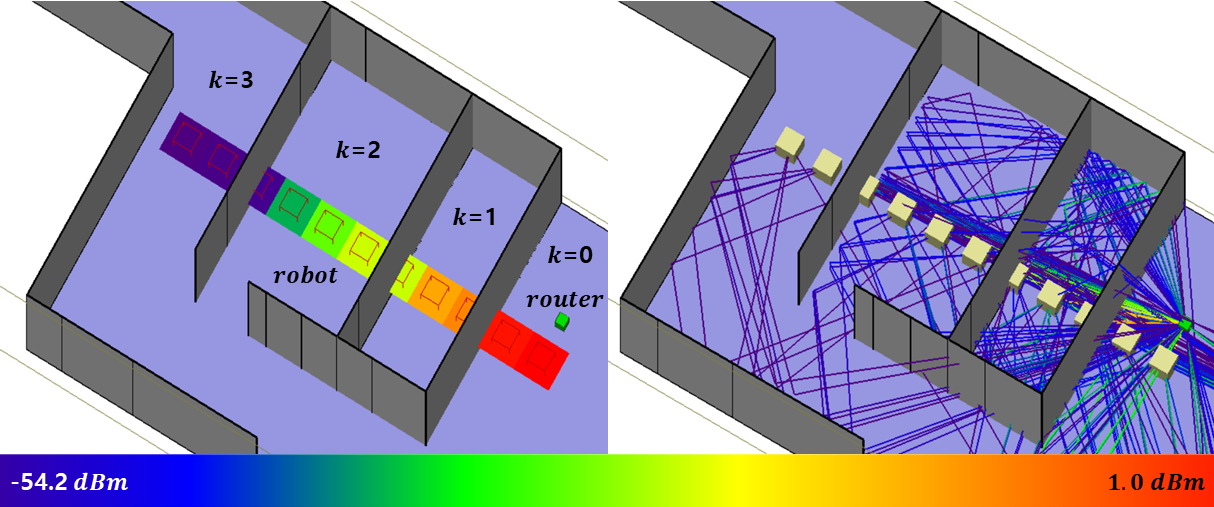}
  \caption{\textcolor{black}{Analysis of received power (left) and propagation paths (right) based on $k$-values using Remcom Wireless InSite \cite{remcom2024}}}
  \label{fig:remcon}
\end{figure} 

Utilizing $K$-Means algorithm in the context of $K$ walls via the algorithm by Fafoutis {\it et al.}, the resulting output is $K+1$ centroids, of which we sort as $C_{0}, C_{1}, ..., C_{K}$. Bounds on the RSSI can then be defined as:

\begin{equation}
    t_{K} = \frac{C_{k-1} + C_{k}}{2}: k\in [1, K]\label{eq:bounds_on_rssi}
\end{equation}

This framework and the guiding equations define the basis of our proposed sparse inverse $k$-visibility algorithm. We conduct experimental trials utilizing this framework.


In real-world experiments, RSSI signals oscillate greatly and exhibit much noise. To account for this noise and thus the fluctuation of the RSSI measurements in hopes of improving accuracy, we apply a sliding window filter as the wearable sensor traverses its trajectory path.


To validate that the $k$-visibility maps in radio propagation, Fig.~\ref{fig:remcon} presents a 3-D ray-tracing simulation conducted using Remcom Wireless Insite \cite{remcom2024}. 
The received power heatmap (left) and the propagation path (right) illustrate a step-wise attenuation pattern consistent with the $k$-visibility formulation. Each additional wall intersection results in a power drop aligning with the empirically derived thresholds used in the RSSI-to-$k$ mapping of Eq.~\eqref{eq:rssi_function}.

\begin{figure}[t]
  \centering
  \includegraphics[width=0.75\linewidth]{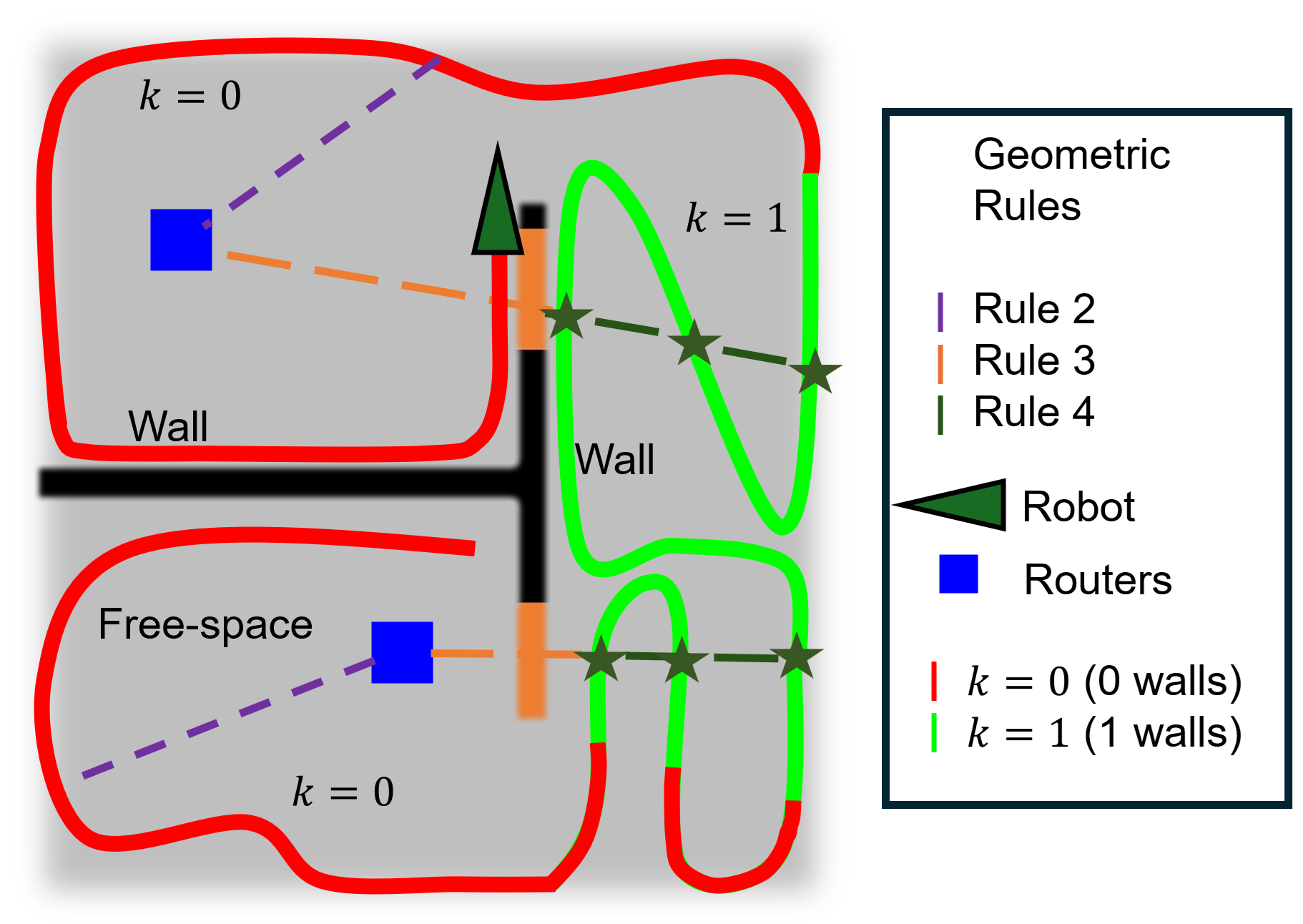}
  \caption{Visual demonstration of the geometric rules for areas with $k=0$ and $k=1$. Rule 1, depicting the robot's trajectory, was omitted from the legend for clarity. 
  }
  \label{fig:rules}
\end{figure}
\subsection{Mapping Free Space}\label{subsec:free}

\color{black}

To map the free space, we employed a set of geometric rules as illustrated in Fig.~\ref{fig:rules}. Initially, we assume that the pixel values of the map are unknown setting to 127. The trajectory is first segmented according to the $k$-values, as detailed in Sec.~\ref{subsec:k}. Subsequently, these rules are applied to identify the free space. Full details are described in Alg.~\ref{alg:sfw_algorithm2_real_time}.


\color{black}


    
    \noindent {\bf Rule 1}: The robot's trajectory is considered as free space unless it encounters an obstacle, detected by contact sensors.
    
    \noindent {\bf Rule 2}: If $k=0$, the pixels on the line segment between the router and the robot are free space.
    
    \noindent {\bf Rule 3}: If $k \geq 1$, there are walls between the router and the robot.
    
    \noindent {\bf Rule 4}: For any line emanating from a router, if the line intersects the robot trajectory at two points with the same $k$-values, the pixel residing on the line that is between the two points belongs to the free space.

    \noindent {\bf Rule 5}: As the $k$-values are determined, the confidence level of each pixel being classified as free space or a wall is updated probabilistically.

\subsection{Mapping Occupied Space}\label{subsec:occupied}

Segmenting the trajectory based on the $k$-value and continuity is the initial step in the algorithm for identifying occupied/free space. The algorithm categorizes trajectory coordinates according to their respective $k$-values.


To identify free space, a ray is drawn from the Router $R$ to all trajectory coordinates with $k$-values equal to $0$. By the definition of $k$-visibility, no walls obstruct these lines; thus, all cells along these $k_0$ rays are classified as free space.


Following this, the outer walls are approximated by drawing a rectangular bounding box around all data points, including trajectory coordinates and cells identified by the rays. This provides an outline of the indoor floor plan, based on the assumption that the robot navigates through all rooms within the map. 

\begin{figure}[b]
  \centering

  \includegraphics[width=.9\linewidth]{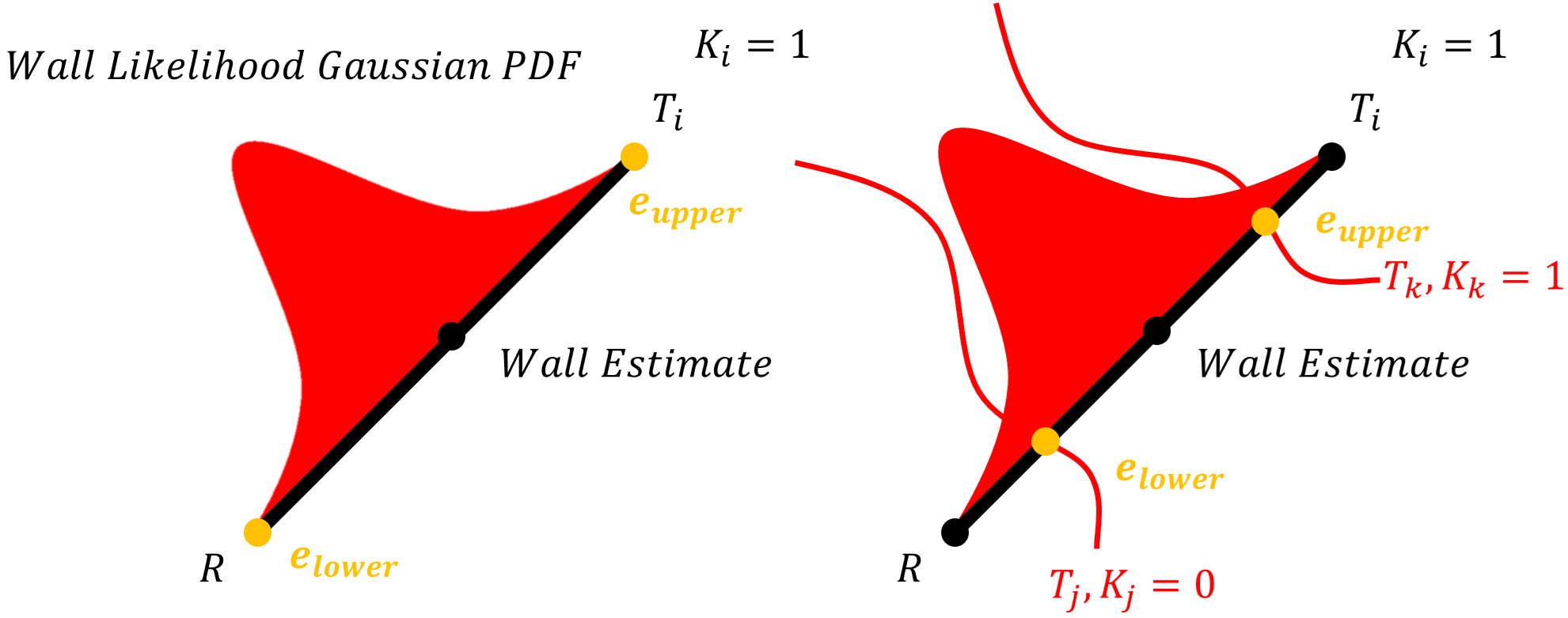}
  \caption{Initial wall estimate along a ray (left). Enhanced wall estimation along a ray after adjustments to the lower and upper endpoints (right).}
  \label{fig:gaussdemo12}
\end{figure}

%
To further identify walls within the map, a probabilistic model is applied based on Geometric $Rule 5$. Consider the ray $\overline{RT_i}$ depicted in Fig.~\ref{fig:gaussdemo12}, where the $k$-values for the trajectory position $K_i = 1$. Given the $k$-visibility, it is known that $\overline{RT_i}$ intersects exactly $K_i$ wall cells, which, in this case, is one wall cell. With no other known information, the most reasonable assumption is that this wall cell is located at the midpoint of $\overline{RT_i}$. The validity of this assumption decreases as the length of the ray increases. 
This probabilistic certainty is represented as:
\begin{equation} \label{eq:intermediate_cell_probability}
    \mu_j = \frac{d_j e^{-(\frac{1}{M})^2}}{L},
\end{equation}

where $\mu_j$ is the probability of the $j^{th}$ cell being a wall, M is the number of intermediate cells along the ray, $L$ is the length of the ray, and $d_j$ is the distance from the $j^{th}$ cell to the ray's midpoint. Intermediate cells are defined as all cells along the ray, excluding the endpoints.


The prediction accuracy for ray $\overline{RT_i}$ in Fig.~\ref{fig:gaussdemo12} can be improved by adjusting the ray endpoints, resulting in a shorter line and fewer intermediate cells. 
Initially, the lower endpoint of a ray, $e_{lower}$, is the router point R, and the upper endpoint, $e_{upper}$, is the trajectory point $T_i$. If the ray intersects other trajectory points, these points serve as updated endpoints. The lower endpoint is updated if the intersecting trajectory point $T_j$ has $K_j = 0$; otherwise, the upper endpoint is updated if~$K_j\geq1$.


\begin{figure}
  \centering

  \includegraphics[width=0.65\linewidth]{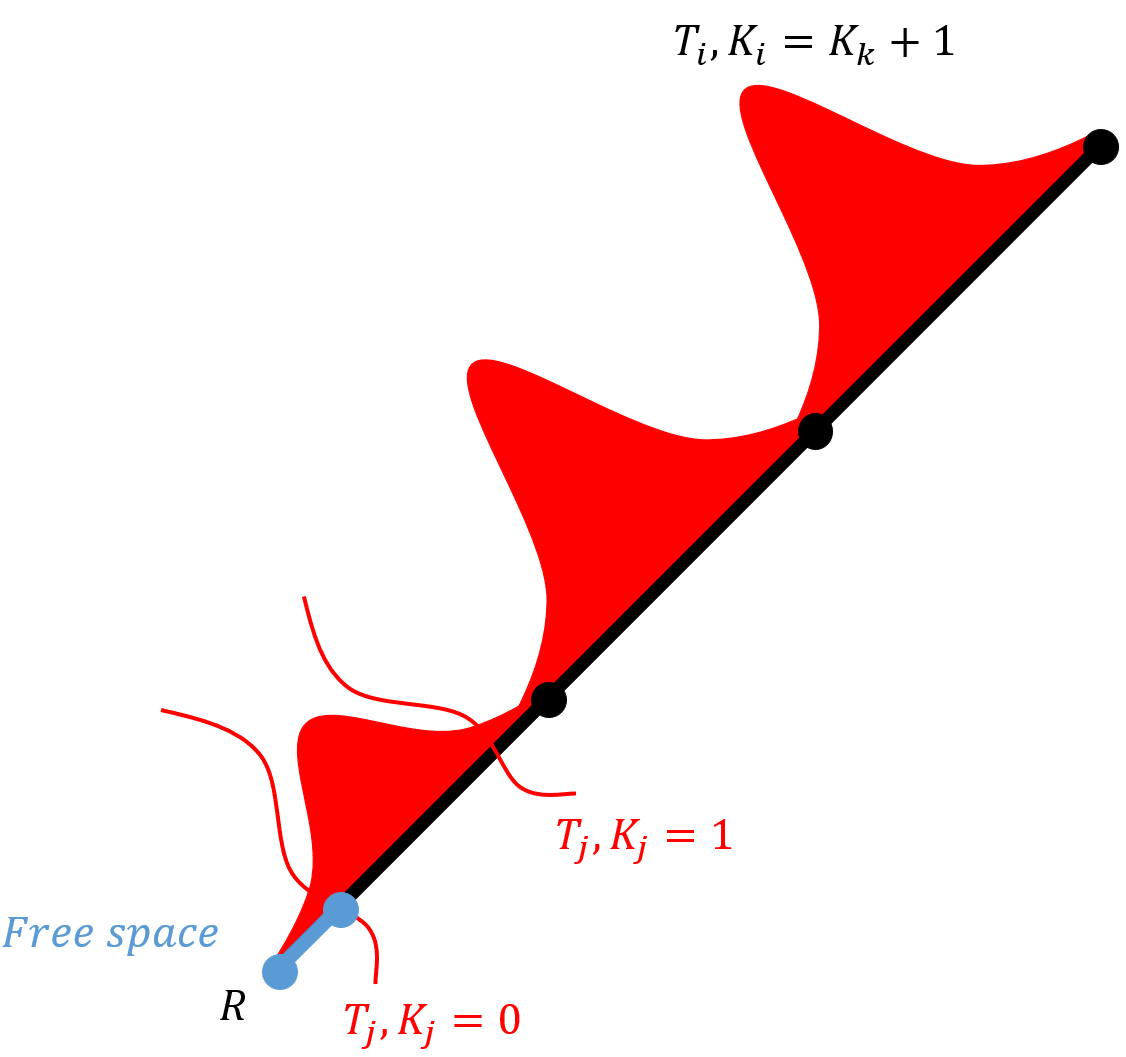}
  \caption{Assigning probability distributions based on the variations in $k$-values across ray subsegments. 
  }
  \label{fig:raysubsegs}
\end{figure}


The approach can be extended to handle cases where $k$-values are greater than 1. For each trajectory coordinate, a ray $\overline{RT_i}$ is drawn, and trajectory intersections are identified as previously described. The ray is then segmented based on these trajectory intersections. The difference $\Delta k$ between the endpoints of each subsegment is calculated, as shown in Fig~\ref{fig:raysubsegs}. If $\Delta k = 0$, meaning that the endpoints have the same $k$-value, the segment is considered as free space. If $\Delta k = 1$, indicating that exactly one wall lies along the subsegment, an unimodal probability distribution is applied, with the highest probability at the midpoint, assuming no additional information is available. For $\Delta k > 1$, multiple walls are along the subsegment, and a multimodal distribution is applied.


When rays are drawn for each cell along the trajectory, it's possible for a cell to be ``seen" from multiple rays. In such cases, the cell would already have an assigned probability from a previous ray, {as calculated by Eq. \eqref{eq:intermediate_cell_probability}} 
The probability for the cell is then updated by integrating the probabilities from both the prior and current rays, weighted by their respective uncertainties. 

\begin{equation} \label{eq:intermediate_cell_probability2}
    \mu = \frac{\sigma_1^2}{\sigma_1^2+\sigma_2^2} \mu_2 +  \frac{\sigma_2^2}{\sigma_1^2+\sigma_2^2} \mu_1
\end{equation}

Here, $\mu$ represents the combined probability, $\mu_1$ and $\mu_2$ are the probabilities assigned by the previous and current rays, respectively, and $\sigma_1$ and $\sigma_2$ are the associated uncertainties of these probabilities.

{These procedures are outlined in Alg.~\ref{alg:sfw_algorithm2_real_time}. The robot's position $(x, y)$ is initially set within its trajectory $T$ (line \ref{ln:ray_init}). The map $M$, initialized to a pixel value of 127, indicating unknown space, corresponding to a probability of occupancy of 0.5 (line \ref{ln:traject_state}). The inputs are the 2D position of the router and sparse $k$-value along the robot trajectory. The output is a 2D occupancy map with free space and potentially occupied cells. The map is updated as the robot explores the environment. The positions of $n$ routers, $(a_1,b_1), (a_2,b_2),..., (a_n,b_n)$, are defined within the map, and the focused router position is initially set to the first router, $(a_1,b_1)$, by default (lines \ref{ln:position_router}). As the robot moves to the next trajectory point $(x_j, y_j)$, it marks the current position $(x_i, y_i)$ as free space with a pixel value of 255, corresponding to a probability of occupancy of 1 (Rule 1) (line \ref{ln:ln6-rule1}).

\begin{algorithm}[t!]
\caption{Multi-Router Free Space Mapping with Incremental K-Means}
\label{alg:sfw_algorithm2_real_time}
\begin{algorithmic}[1]
\STATE \textbf{Inputs:} 
\STATE \quad Reference point \( R \) (e.g., WiFi router position) \label{ln:input3}
\STATE \quad sparse $k$-value along the robot trajectory \label{ln:input4}

\STATE \textbf{Outputs:}
\STATE \quad A map \( M \) that classifies regions as free space or walls \label{ln:output1}
\STATE \textbf{initialize:} Robot position $(x, y) \in T$, \label{ln:ray_init}
\STATE trajectory $T \in M$, map $M_{ij} \in \{\mathbb{R}^2\}$ $\forall i,j$, $M_{ij} \gets 127$, \label{ln:traject_state}
\STATE router positions $(a_1, b_1), (a_2, b_2),..., (a_n, b_n) \in M$, focus $(a,b) \gets (a_1,b_1)$. \label{ln:position_router}
\STATE Empty dataset $\mathcal{D} \gets \{\}$, set $K$ for k-means clustering.

\WHILE{Robot moves to $(x,y)_i \gets (x,y)_{i+1}$}
    \STATE $M_{(x,y)_i} \gets 0$ (free space) [Rule 1]. \label{ln:ln6-rule1}
    \STATE $\mathcal{D} \gets \mathcal{D} \cup \{P_{RSSI}\}$.
    \FORALL{$(a_k, b_k) \in \{(a_1, b_1), ..., (a_n, b_n)\}$} 
        \STATE $R_k \gets P_{RSSI} \text{ from } (t_x, t_y) \text{ to } (a_k, b_k)$. \label{ln: ln9}
        \IF{$R_{k-1} < R_k$ and $k > 1$}
            \STATE $(a,b) \gets (a_k, b_k)$. \label{ln: focus_update}
        \ENDIF
    \ENDFOR

    \FORALL{$(t_x, t_y) \in M$ from $(x, y)$ to $(a, b)$} \label{ln: for_loop_alg}
        \STATE $L_i \gets P_{RSSI} \text{ from } (t_x, t_y)$.
        \STATE Run k-means on $\mathcal{D}$, update centroids $C_0, C_1, ..., C_K$.
        \STATE Update thresholds: $t_k = \frac{C_{k-1} + C_k}{2}$.
        \STATE $K_i \gets f(P_{RSSI})$ based on updated $t_k$. \label{ln: k_update}
        
        \IF{$K_i = 0$}
            \STATE $L_i \gets L_i + \sigma$ (free space) [Rule 2]. \label{ln: rule_2_multi}
        \ELSIF{$K_i \geq 1$}
            \STATE$L_i \gets L_i - \sigma$ (wall) [Rule 3]. \label{ln: rule_3_multi}
        \ENDIF
        
        \IF{$\exists c,d \in \mathbb{R} \text{ such that } K_c = K_d$}
            \FORALL{$L_i \text{ from } L_c \text{ to } L_d$}
                \STATE$L_i \gets L_i + \sigma$ (free space) [Rule 4]. \label{ln: rule_4_multi}
            \ENDFOR
        \ENDIF
        \STATE Update $M$ with $P_{RSSI} \text{ of } (t_x, t_y) \gets L_i$. \label{ln: update_map}
    \ENDFOR

    \STATE Periodically re-run k-means and update thresholds after $n$ points.
\ENDWHILE
\end{algorithmic}
\end{algorithm}
\begin{figure}[h]
  \centering
  \includegraphics[width=1\linewidth]{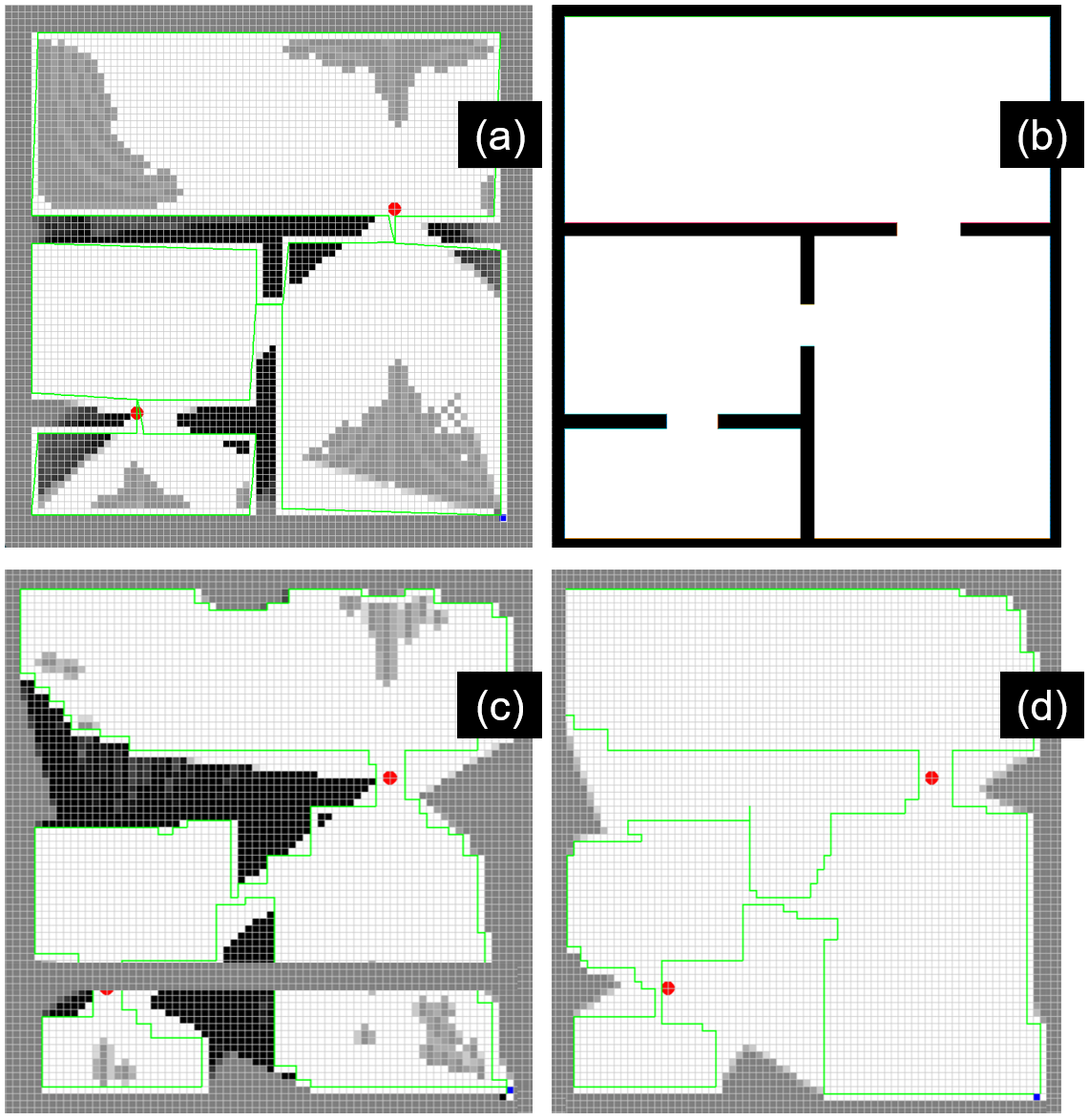}
  \caption{Simulated results utilizing the proposed $k$-visibility algorithm. (a) Free space and occupied cells, with the robot's trajectory depicted in green. (b) The ground-truth map~\cite{"li2019houseexpo"}. 
  (c) 2D map generated for the same environment 
  with another trajectory when the $k$-visibility is considered. (d) The new trajectory builds a map that marks most cells free when full $k$-visibility is not considered. In this case, any cell residing on the rays between a router and a trajectory point is considered free. This is equivalent to line-of-sight of $k$-visibility with $k=0$. The map is less reliable in this case.
  }
  \label{fig:simgroundtruth}
\end{figure}

For each router $(a_k,b_k)$, the algorithm computes the RSSI value $R_k$ from the current robot position $(x_i, y_i)$ to the router (line \ref{ln: ln9}). If the new RSSI value $R_k$ is stronger than the previously recorded value from another router, the focused router is updated to the current one, $(a_k, b_k)$ (lines \ref{ln: focus_update}). The algorithm then iterates through all points $(t_x, t_y)$ on the line segment between the robot's current position $(x_i, y_i)$ and the focused router $(a_k, b_k)$ (line \ref{ln: for_loop_alg}). For each point, it calculates the $k$-value using a probabilistic model based on the signal strength and the number of obstacles between the robot and the router (lines \ref{ln: k_update}).

If the computed $k$-value $K_i = 0$, meaning there are no obstacles, the probability of the space being free is increased by $\sigma$, and the pixel value is moved closer to 255 (Rule 2) (lines \ref{ln: rule_2_multi}). Conversely, if $K_i \geq 1$, indicating that there are obstacles, the probability of the space being occupied is increased, reducing the pixel value closer to 0 (Rule 3) (lines \ref{ln: rule_3_multi}).

If two points $c$ and $d$ along the ray from the router have the same $k$-value ($K_c = K_d$), the algorithm assumes that the space between them is free and updates the probability of all points along the line between $c$ and $d$, increasing their pixel values closer to 255 (lines \ref{ln: rule_4_multi}). Throughout this process, the map $M$ is continuously updated with RSSI values, gradually constructing the environment map as the robot traverses its trajectory (lines \ref{ln: update_map}).}

\textcolor{black}{In Alg.~\ref{alg:sfw_algorithm2_real_time}, we adopt the standard occupancy-grid practice of fixed evidence updates (log-odds steps) for free and occupied, which is equivalent to using a constant $\sigma$ per update event rather than re-estimating a distance-varying $\sigma(d)$ at every cell. This choice mirrors conventional inverse-measurement-model update~\cite{thrun2005probabilistic} and retains the Bayesian interpretation at the grid-cell level while still being consistent with the probabilistic model in Eq.~\eqref{eq:intermediate_cell_probability} and Eq.~\eqref{eq:intermediate_cell_probability2}.}

\section{Experimental Results} \label{sec:exp}
\color{black}
This section presents the implementation of sparse inverse $k$-visibility under simulated and real-world experiments to validate its robustness. We also test the improvement from single-router to multi-router cases.

In the simulated environment, as shown in Fig.~\ref{fig:simgroundtruth}, a robot travels along an ideal
trajectory, {defined as a trajectory path that follows along the walls and explores all walls of each room. This ensures detailed mapping of the environment's boundaries.} The floorplan map was provided by the HouseExpo dataset~\cite{"li2019houseexpo"}. As the robot travels around the room, every trajectory coordinate is associated with a unique $k$-value based on its current and relative position to the router. From the simulated result, free space and walls are estimated with great accuracy compared to the ground truth when given the optimal path. {Fig.~\ref{fig:simgroundtruth} presents the results for a slightly different trajectory. In this figure, the significance of $k$-visibility is shown.
Fig.~\ref{fig:simgroundtruth} (left bottom) shows the map developed utilizing the proposed sparse inverse k-visibility algorithm. The free space and location of obstacles are indicated with white and dark cells. The darker a cell is, the more likely it is occupied. Fig.~\ref{fig:simgroundtruth} (right bottom) shows the maps ignoring the $k$-visibility and only utilizing the line-of-sight between the routers and the robots. This is equivalent to executing only Rule 1 and Rule 2. In this case, most of the cells are considered free.}

\begin{figure}[ht!]
    \centering
    \includegraphics[width=0.8\columnwidth]{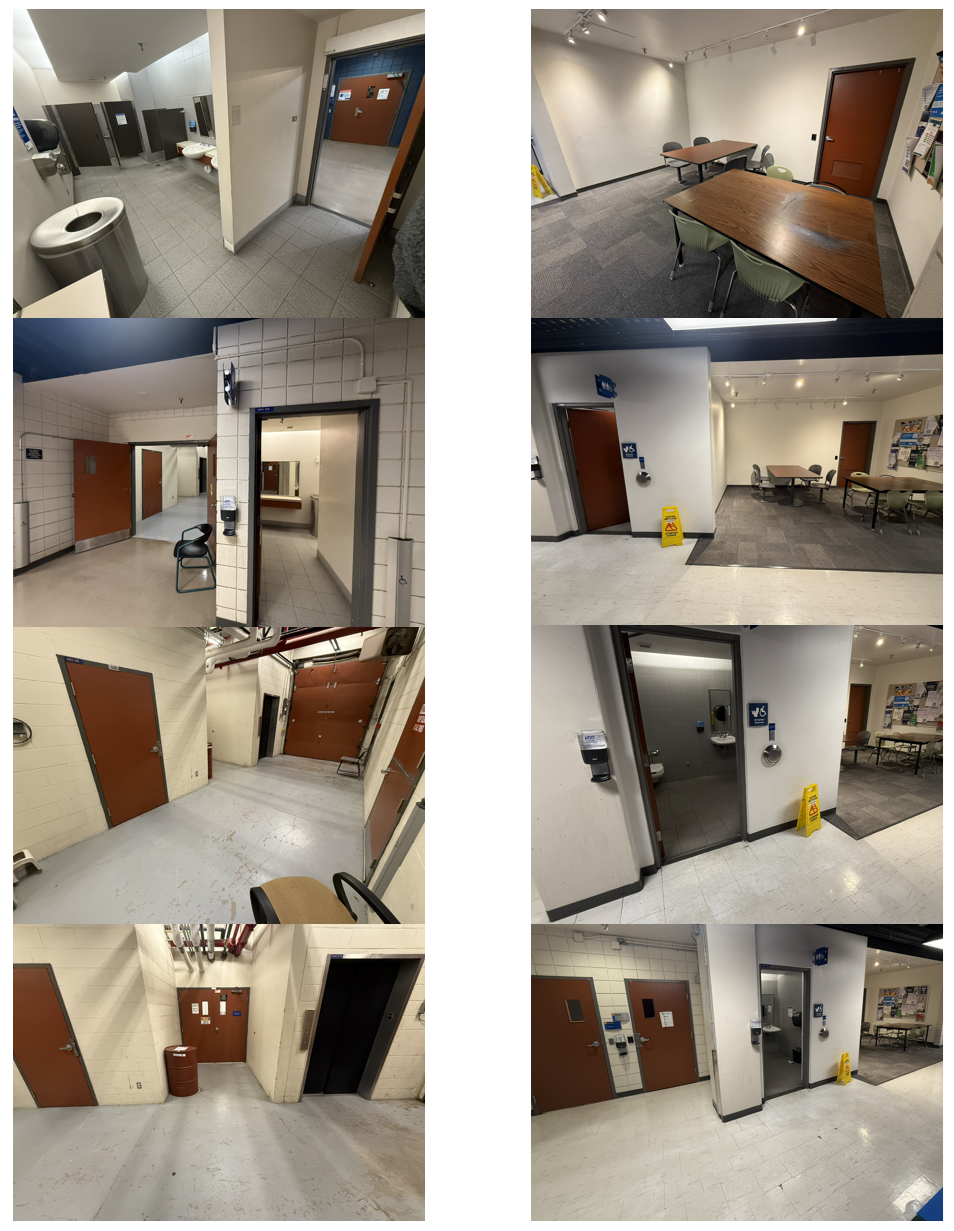}
    \caption{\textcolor{black}{Representative RGB images of the real-world environments used for SfW experiments. The left four images correspond to Exp.~3 and 5, and the right four images correspond to Exp.~4 and 6. These environments include corridors, doorways, cluttered areas, restrooms, and metallic/reflective surfaces, illustrating the environmental complexity and practical challenges encountered during WiFi-based mapping.}}
    \label{fig:env_rgb}
\end{figure}

For the real-world experiments, the LiDAR map using a $360$ Laser Distance sensor (LDS-01) was used as ground-truth data to validate our algorithms on a 2D occupancy grid map. The routers were positioned on the ground floor rather than mounted on walls or ceilings for consistency. 
\textcolor{black}{To provide additional visual context of the experimental settings and to highlight the environmental complexity (e.g., reflective surfaces, narrow corridors, and cluttered areas), representative RGB images of the environments are shown in Fig.~\ref{fig:env_rgb}. These environments introduce realistic multipath effects and attenuation challenges that directly impact RSSI-based mapping performance.}

%
%
Our algorithm is heavily dependent on the optimality and quality of the robot trajectory to estimate the map. The {odometry} data from the TurtleBot3 was used for the $(x,y)$ coordinate of the robot at all points along the trajectory, and RSSI signal strengths were measured at each position. A total of four different trajectories were tested under different environments, including some challenging environments with complex floor plans. In all cases, we assumed that the router's location is known. \textcolor{black}{When unknown, they may be estimated with RF-based localization: RSSI trilateration/multilateration or fingerprinting~\cite{han2009access, koo2010localizing, xia2017indoor, shang2022overview}, but this is a challenging task in multipath-rich settings and is outside the scope of this work.} 
The robot's trajectory was set to explore all walls within the floor plan.

\subsection{Evaluation Metrics}

\color{black}
Prior to any metric evaluation, the RSSI-generated map must be aligned and superposed along the ground-truth LiDAR map, both discretized by an aligned occupancy grid map. Each grid element on the grid map represents a small-element subset of the entire $M\times N$ map, which is used for element-wise comparison.
Here, we will present the evaluation metrics used in measuring the experimental accuracy after alignment.



\noindent {{\bf k-value accuracy Percentage}: We use the wall prediction model proposed by Fafoutis {\it et al.}~\cite{fafoutis2015rssi}, as discussed further in Sec.~\ref{subsec:k} to obtain the $k$-values from the acquired RSSI values which uses a $K$-Means approach. The RSSI-generated map is then estimated by Alg.~\ref{alg:sfw_algorithm2_real_time}. From the ground-truth map, we then construct expected $k$-values based on the positioning of the routers and the wall obstacles.}

Ultimately, the $k\text{-}value_{accuracy}$ is an element-by-element comparison of the estimate $k$-values from the estimate RSSI-generated map and the ground-truth $k$-values from the ground-truth map. If the estimated $k$-value is the same as the expected on the ground truth, it is deemed predicted True and, 
%
%
%
{\begin{equation}
    k\text{-}value_{True}=k\text{-}value_{True}+1\label{eq:increment_rssi},    
\end{equation}
\noindent{with no change if the prediction is not correct, deemed predicted False. After all $M \times N$ elements, the k\text{-}value accuracy is calculated by,}
\begin{equation}
    {k\text{-}value}_{accuracy}= \frac{k\text{-}value_{True}}{M\times N} \times 100 \%. \label{eq:rssi_accuracy}
\end{equation}}
\noindent{\bf IOU Score}: A mask $Z$ on each coordinate $(i,j)\in M\times N$ first needs to be applied on the RSSI-estimate $RSSIE$ map to exclude any estimates that are neither considered free-space nor a wall:
\begin{equation}
Z(i, j) = 
\begin{cases} 
1 & \text{if } RSSIE(i, j) \neq 127 \\
  & \quad \text{and } GT(i, j) \neq 127 \\
0 & \text{otherwise}
\end{cases}.
\end{equation}\label{eq:mask_for_rssi_accuracy}
\noindent where $GT$ is ground-truth map of $M\times N$.
Then, the mask $Z$ is applied to filter out the uncertain parts by an intersection:
\begin{equation}
I(i, j) = 
\begin{cases} 
1 & \text{if } RSSIE(i, j) \neq 0
   \text{ and } GT(i, j) \neq 0 \\
  & \quad \text{and } Z(i, j) = 1 \\
0 & \text{otherwise}
\end{cases}.
\end{equation}\label{eq:intersection_with_mask}
Next, we account for the entire map by a union $U$ where the element is certain (free-space or wall):
\begin{equation}
U(i, j) = 
\begin{cases} 
1 & \text{if } (RSSIE(i, j) \neq 0  \text{ or } GT(i, j) \neq 0) \\
  & \quad \text{and } Z(i, j) = 1 \\
0 & \text{otherwise}
\end{cases}.
\end{equation}
\noindent The Intersection Over Union (IOU) Score, $IOU_{score}$, is then calculated by
{
\begin{equation}
IOU_{score} = \frac{\sum_{i,j} I(i, j)}{\sum_{i,j} U(i, j)}.
\end{equation}
}

\noindent {{\bf MSE Score}:} {First, the difference $D_{i,j}$ is taken element-wise between $RSSIE$ and $GT$ and then squared $DS$:}
{
\begin{equation}
D(i,j) = RSSIE(i,j) - GT(i,j),
\label{eq:difference_rssi_gt}
\end{equation}
\begin{equation}
DS(i,j) = (D(i,j))^2 = (RSSIE(i,j) - GT(i,j))^2.
\label{eq:squared_difference}
\end{equation}
}
{Next, the Mean Squared Error (MSE) Score, $MSE_{score}$, the mean of the square differences $DS$, is calculated by}
{
\begin{equation}
\begin{aligned}
MSE_{score} &= \frac{1}{N \times M} \sum_{i,j} DS(i,j) \\
&= \frac{1}{N \times M} \sum_{i,j} (RSSIE(i,j) - GT(i,j))^2.
\end{aligned}
\label{eq:mse_score}
\end{equation}
{Note, the unit of MSE Score is in Square Pixel (${px}^2$).}
}

\noindent Also, the parameters referred to are the propagation model parameters, and these were first addressed in Sec.~\ref{subsec:k} and Sec.~\ref{subsec:occupied}. The bounds on the RSSI parameter $t_k$ (in Eq.~\eqref{eq:bounds_on_rssi} of our work) in the real-world experiments are found using the centroids provided by the K-Means algorithm on the aligned RSSI-generated map and superposed along the ground-truth LiDAR map for scaling consistency. This clustering provides centroids that serve as thresholds to differentiate different levels of visibility ($k$-values), ensuring consistent scale alignment with the LiDAR-based ground truth. The number of clusters used in the experiment is three as $k$-value of 0, 1, and 2 were defined. 

\begin{figure*}[t]
  \includegraphics[width=\textwidth]{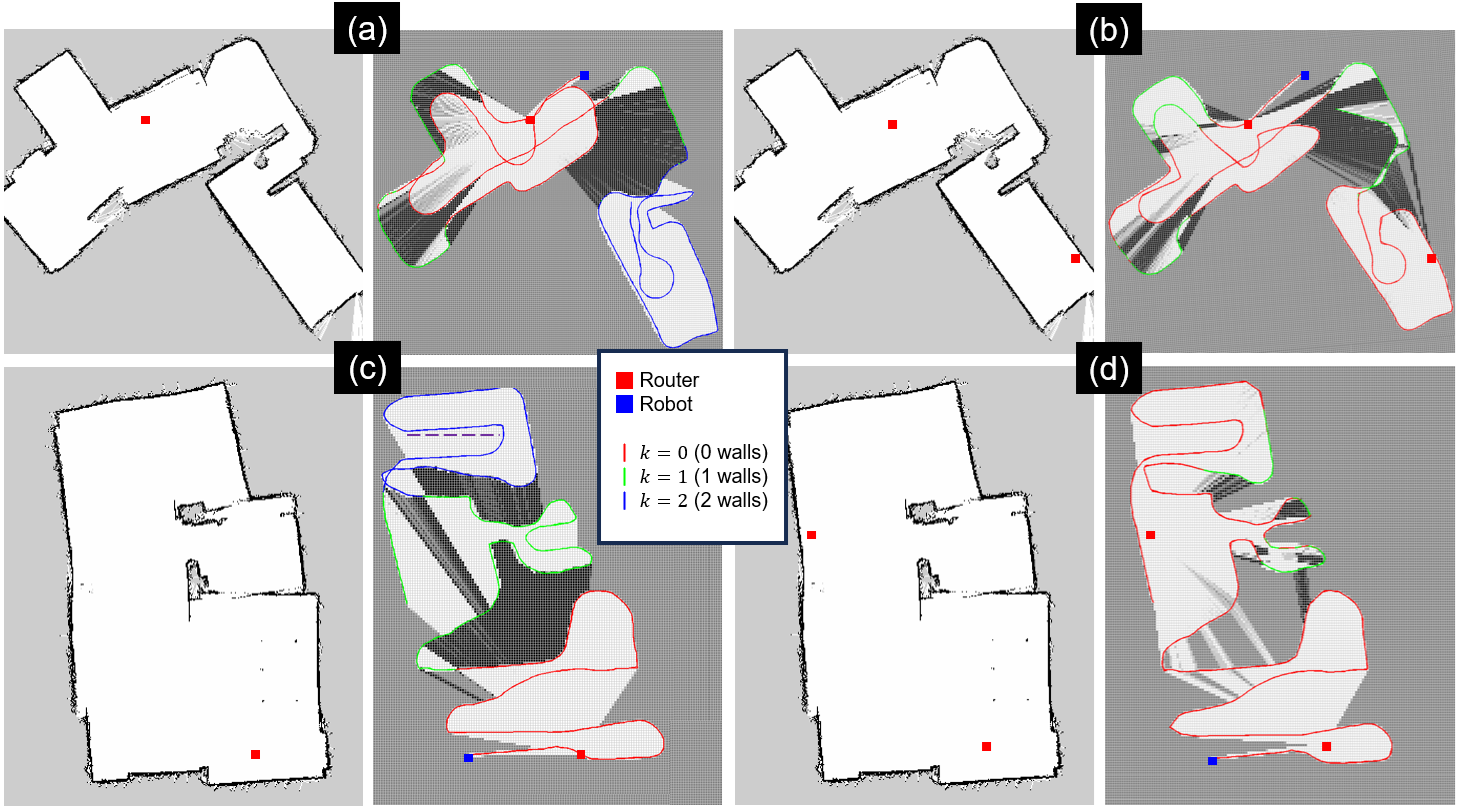}
  \caption{\textcolor{black}{Real-world experimental results for single-router and multi-router SfW mapping. 
  (a) and (b) are for the same environment, but with one and two routers, respectively. The same applies to (c) and (d).
  Our proposed method with RSSI-based mapping detects most of the free space and potential walls with $k\geq 1$. 
  The multi-router case has improved accuracy in estimating its map. Note: For each pair from (a)-(d), the left shows the ground truth obtained from LiDAR, while the right shows the SfW mapping.} 
  }
  \label{fig:all_4_exps} 
\end{figure*}

In Eq.~\eqref{eq:intermediate_cell_probability}, the probabilistic certainty $\mu_j$ of the $j$-th cell being a wall in our LOS point along the robot's trajectory is handled within the algorithm rather than shown in each instance, given the large number of trajectories and overlapping segments encountered. 
Similarly, Eq.~\eqref{eq:intermediate_cell_probability2}, where the combined probability of the cell by multi-ray intersection updating $\mu$ is done in the algorithm to reflect superposed probabilities from intersecting rays, including $\mu_1$ and $\mu_2$, without needing to detail every recalculation.

For real-world adjustments, the $\sigma$ parameter associated with free space probabilities along each pixel's LOS path was specifically tuned. This $\sigma$ value, assumed to follow a Gaussian distribution based on RSSI fluctuations near each pixel, was refined to closely align the RSSI-derived map with the ground truth obtained from LiDAR. Different $\sigma$ values were tested based on data volume, with larger datasets providing more overlapping trajectories. With more overlap, incremental updates to $\mu$ values can be smaller, balancing the density of trajectory data with accurate map representation. This introduces a trade-off: dense data allows smaller $\sigma$ adjustments, while sparse data benefits from slightly larger adjustments to maintain consistency across the map. In Fig.~\ref{fig:all_4_exps}, the model was tested with a $\sigma$ pixel value of 10.

\color{black}


\subsection{Discussion: Ignoring $k$-visibility}

\begin{figure}[t]
  \centering
  \includegraphics[width=1\linewidth]{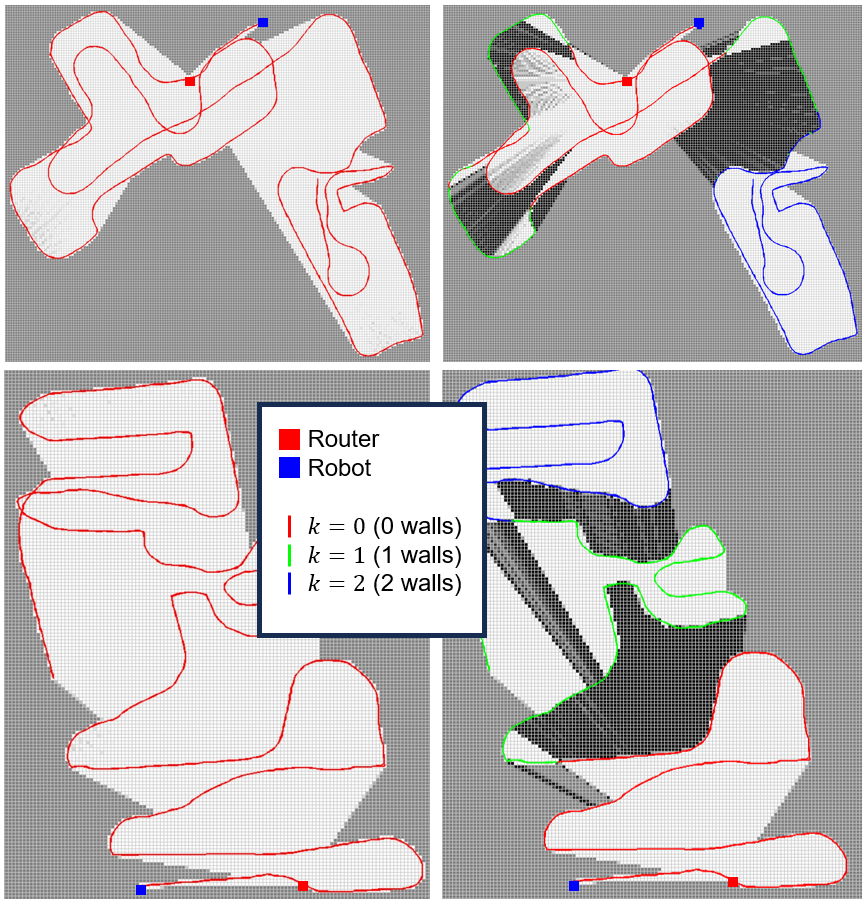}
  \caption{\textcolor{black}{Real-world results of the comparison of mapping results between LOS alone (left top and bottom) and the inverse $k$-visibility algorithm (right top and bottom). LOS alone is insufficient when wanting to also have a measure of spatial reliability on the mapping prediction.}}
  \label{fig:los-vs-k}
\end{figure}

Using Line-of-Sight (LOS) alone will not provide enough information to give a very fine and detailed prediction of the entire map. This is treating all $k$-values as $k=0$, which would result in mapping the entire space as free. In this case, only WiFi signals are used, resembling applications such as a person walking in an environment to map it using only a WiFi receiver, or a robot with only a WiFi receiver.  The inverse $k$-visibility algorithm provides a more sophisticated approach by distinguishing between different visibility values. In using a set of different $k$-values, the detail of the mapping prediction is provided, which would be a strong metric or basis to use in knowing the trustworthiness of the prediction. If there were no varying $k$-value with all $k=0$, then it would be hard to determine which parts of the predicted map are accurate enough, especially in robotic applications where mapping accuracy is essential. The comparison of doing mapping with LOS alone compared with the algorithm is provided in Fig.~\ref{fig:los-vs-k}. All in all, this will ensure a more accurate representation of the space, particularly in environments where walls and obstacles exist.

\subsection{Quantitative Results}

\color{black}

{For all experiments, CPU 12th Gen Intel(R) Core(TM) i7-12650H was used to evaluate different methods. The experiments were performed in an indoor environment with 
walls and closed layouts, chosen specifically to ensure that direct-path signals dominate the received measurements. 
\textcolor{black}{Note that no existing WiFi-only method outputs a directly comparable occupancy map under the same sensing assumptions and evaluation protocol. Prior works are primarily aimed at RSSI heat map generation~\cite{zou2020adversarial} or robot/access-point localization rather than explicit occupancy mapping~\cite{ferris2007wifi, arun2022p2slam}. Approaches that reconstruct occupancy from wireless measurements employ different sensing hardware or experimental setups, such as directional antennas~\cite{gonzalez2013cooperative}. Because these differences affect both the available information and the evaluation conditions, we report internal metrics (IoU, MSE) computed on aligned grids against LiDAR ground truth within our test environments rather than a side-by-side baseline.} Experiments 
include comparison by mapping area, number of data points, $k$-value prediction True/False setting, $k$-value accuracy percentage, IOU, and MSE scores. Further analysis was conducted to evaluate our proposed method, as shown in Table~\ref{tb:table1}. 
{Experiments 5-6 show improved performance with fewer data points. This is because their experiments utilize two routers instead of one. With two routers, more data is collected simultaneously, and the most optimized data is selected from various cases. In contrast, using a single router provides only one piece of information at a time, causing less efficient data collection.}

In the first experiment, a total of 3159 data points were collected and showed 84.93\% $k$-value prediction accuracy. Here, the IOU 
score was calculated at 0.8531, comparing each corresponding pixel from the ground truth and the {RSSI-generated} map. The MSE score was 1.7770 {${px}^2$}, showing high similarity with the ground truth.

Similarly, the second experiment showed an 82.79\% accuracy rate out of 4713 data points. The area of the map was more than twice of the first experiment. The IOU score and the MSE score were 0.9028 {${px}^2$} and 17.4238 {${px}^2$} respectively. 


\begin{table*}[ht]
\centering
\caption{Accuracy and map similarity comparison between LiDAR (Ground Truth) and RSSI generated map. For brevity, $\uparrow$ means a larger result is better, and $\downarrow$ means a lower result is better.}
\label{tb:table1}
\footnotesize
\setlength{\tabcolsep}{4.0pt}
\renewcommand{\arraystretch}{1.7}
\begin{tabular}{ccccccc}
\hline
& \textbf{Exp.1} & \textbf{Exp.2} & \textbf{Exp.3} & \textbf{Exp.4} & \textbf{Exp.5} & \textbf{Exp.6} \\
\hline
Area of Map & \textbf{15.7~m\textsuperscript{2}} & \textbf{34.2~m\textsuperscript{2}} & \textbf{42.5~m\textsuperscript{2}} & \textbf{51.5~m\textsuperscript{2}} & \textbf{42.5~m\textsuperscript{2}} & \textbf{51.5~m\textsuperscript{2}} \\ 
\hline
\#Router & 1 & 1 & 1 & 1 & 2 & 2 \\ 
\hline
\#Data points & 3159 & 4713 & 6001 & 1803 & 492 & 393 \\ 
\hline
Total computation time & 32.94 s & 34.05 s & 46.18 s & 30.75 s & 3.04 s & 2.88 s \\ 
\hline
Time per data & 5.21 ms & 3.61 ms & 3.85 ms & 8.53 ms & 3.09 ms & 3.66 ms \\ 
\hline
$k$-value Prediction True & 2683 & 3902 & 5120 & 1698 & 471 & 379 \\ 
\hline
$k$-value Prediction False & 476 & 811 & 881 & 105 & 21 & 14 \\ 
\hline
$k$-value Accuracy \% $\uparrow$ & 84.93 & 82.79 & 85.32 & 94.18 & 95.73 & 96.44 \\ 
\hline
IOU Score $\uparrow$ & 0.8531 & 0.9028 & 0.8321 & 0.9349 & 0.8364 & 0.9627 \\ 
\hline
MSE Score $\downarrow$ & 1.7770 & 17.4238 & 9.0783 & 12.1403 & 1.6163 & 13.5331 \\ 
\hline
\end{tabular}
\end{table*}

\subsection{Single-Router to Multi-Router Evaluations}

Experiments 3-6 test the proposed algorithm for multi-router cases, where it is expected that additional router sensors would greatly improve the accuracy of the map as the WiFi signal strength is superposed and strengthened. We will start with details on single-router, then introduce more.

Starting with the single-router case, the third experiment in Fig.~\ref{fig:all_4_exps}-(a) covered a trajectory dealing with $6001$ data points in an area of 42.5$~m^2$ and showcased high $k$-value accuracy of 85.32\%. In this experiment, we placed the router in a room with higher-density walls and corridors, and our algorithm can still maintain high robustness in its map. The large black spaces in our generated RSSI-based map, which are considered wall space, were, in fact, higher-density walls showcasing the accuracy of our method. Due to the properties of our algorithm, large black spaces can be further classified as either free space or a wall with more trajectory data and RSSI 
data as the robot travels in that specific area. 


With the multi-router case for the fifth experiment in Fig.~\ref{fig:all_4_exps}-(b), the $k$-value accuracy greatly improves, going from 85.32\% to 95.73\% with significantly fewer data points. The second router is leveraged in the $k$ region where prior single-router cases struggled. As our algorithm makes each router to only prioritize the closest and strongest signal, the $k=0$ regime is effectively identified with minor false prediction. Details on the multi-router case for the third experiment are stated as experiment~5 in Table~\ref{tb:table1}. 

Next, with the single-router case for the fourth experiment in Fig.~\ref{fig:all_4_exps}-(c), dealing with a map area of 51.5$~m^2$ was the largest map area space covered out of all four experiments. This case was unique, where the router was placed at one end to eliminate any effects that the distance loss might have caused. As expected, as $k$ increases, the accuracy of our RSSI-based method might deteriorate. This is certain for conventional mapping methods as well. As shown, for $k=0$, the space generated by the RSSI-based approach is close to the ground truth. As the trajectory moves in the $k=1$ space, the map generated is still highly accurate, with two large black regions representing the wall spaces in our ground-truth map. Going to $k=2$, the expected free space is mapped. This experiment shows that in large area spaces where the router is not in close proximity to the robot but instead very far, our RSSI-based approach still maintains robustness in mapping, with an accuracy of 94.18\%  to the real map.


With the multi-router case for the sixth experiment in Fig~\ref{fig:all_4_exps}-(d), again, the RSSI map accuracy improves, this time going from 94.18\% to 96.44\%. The increased improvement is not as great as the third experiment, only with a 2\% increase, but this is due to the larger area of the fourth experiment. In comparing the single-router with the multi-router case, it is clear that in regions where the single-router map estimate fails, the multi-router tracks more accurately the free space area, whilst circumventing the false positives of walls. Experiment~6 accurately differentiates the free space/potential walls, which were poorly detected in the $k=2$ region, from the fourth experiment. From the evaluation metric, the computational time per RSSI was calculated to be between 3ms to 8ms. The exact number can vary depending on the hardware capacity. However, it shows that the algorithm can process the RSSI in real-time. \color{black}Details on the multi-router case for the third experiment are stated as experiment~6 in Table~\ref{tb:table1}.

\textcolor{black}{In our implementation, the ``focused router" is chosen as the one with the strongest RSSI, and updates are applied along that router's ray only. While this simplifies updates and improves $k=0$ detection, it can leave bridging regions between routers unobserved when the trajectory remains within one router's dominance zone if the trajectory does not cross-link the zones. We recommend targeting that band with active exploration (frontier/information-gain moves and short cross-connecting passes). See Sec. \ref{sec:con} for details and references.}

\subsection{Discussion}

While RSSI data points ranging between 393 to 6001 were collected from four experiments, all experiments have shown good estimates of the $k$ values above 82\%. These four experiments show a robust prediction capability of the RSSI signals across various environments in its high $k$-value accuracy percentage, its near 1 IOU score, and its relatively low MSE score, shown in Table~\ref{tb:table1}.

The $k$-value accuracy percentage was optimized higher in experiments 4-6 in larger map areas compared to the prior experiments, by the positioning of the routers and the careful traversing of the robot trajectory to optimize RSSI signal strength from the routers. This implies that methodically planning the trajectory in the map estimate is essential.
 
The IOU score, which measures the overlap between predicted free space and the ground truth, showed little variability for all experiments, ranging from 0.8-0.96, where a near 1 IOU Score indicates a good overlap between the ground truth and our algorithms.

Though the $k-value$ accuracy percentage and IOU score showed great accuracy and robustness in the map estimate over all experiments, the MSE score had significant variability. Experiments 2-4 and 6 shown in Table~\ref{tb:table1} had larger MSE scores, but given the large area spaces, these MSE scores are within reason and alleviated when going from single-router to multi-router cases, as shown by the significant drop in MSE score of single-router experiment~3 with MSE score of 9.0783 {${px}^2$} to multi-router experiment~5 with MSE score of 1.6163 {${px}^2$}.

\textcolor{black}{\subsection{Sensitivity analysis}}

\textcolor{black}{In the same simulation setup as Fig.~\ref{fig:simgroundtruth}, we evaluated robustness to two uncertainties: (i) router-coordinate noise, modeled as zero-mean Gaussian perturbations with standard deviation $\sigma\in\{0,1,2,5,10,20\}$ px (1 px $\approx$ 0.025 m) and (ii) odometry drift, modeled as zero-mean Gaussian increments that accumulate over time with per-step standard deviation $\sigma\in\{0,0.1,0.2,0.5,1.0,2.0\}$ px. The smaller per-step values for odometry reflect realistic compounding drift in practice. For router coordinate perturbations, performance remained essentially stable (Table~\ref{tab:router_sensitivity}), indicating overall robustness to modest placement errors. However, this robustness is conditional on line-of-sight (LOS). When router trajectory rays remain predominantly clear (e.g., centrally placed routers with broad coverage), the geometric cues are informative and performance is stable. In contrast, unfavorable placements (e.g., corner locations or frequent occlusions) reduce the number of useful sightlines, which degrades the spatial informativeness of the measurements and thus tends to decrease IoU and increase MSE.}
\textcolor{black}{On the other hand, odometry with accumulating noise degrades map quality sharply: small per-step perturbations compound over time, and for noise standard deviation $\geq 0.5$, the drift overwhelms wall-based constraints, leading to failure to complete the map. These results underscore the need for re-localization or sensor fusion to bound long-term error (Table~\ref{tab:odometry_sensitivity}). Visualizations for router-coordinate noise and odometry noise can be found in Fig.~\ref{fig:router_noise} and Fig.~\ref{fig:odom_noise}.}

\setcounter{table}{1}
\begin{table}[th]
\centering
\caption{\textcolor{black}{Router coordinate sensitivity (Gaussian perturbation of perceived router locations).}}
\label{tab:router_sensitivity}
{\color{black}
\setlength{\tabcolsep}{18pt}
\renewcommand{\arraystretch}{1.05}
\begin{tabular}{@{}ccc@{}}
\toprule
\textbf{Noise Std (px)} & \textbf{IoU $\uparrow$} & \textbf{MSE $\downarrow$} \\
\midrule
0  & 0.9556 & 7.45 \\
1  & 0.9554 & 7.49 \\
2  & 0.9552 & 7.48 \\
5  & 0.9549 & 7.33 \\
10 & 0.9542 & 7.27 \\
20 & 0.9538 & 7.91 \\
\bottomrule
\end{tabular}
} 
\end{table}

\begin{table}[th]
\centering
\caption{\textcolor{black}{Odometry sensitivity under cumulative drift. \emph{Fail} indicates the run did not complete successfully (drift caused divergence). }}
\label{tab:odometry_sensitivity}
{\color{black}
\setlength{\tabcolsep}{18pt}
\renewcommand{\arraystretch}{1.05}
\begin{tabular}{@{}ccc@{}}
\toprule
\textbf{Noise Std (px)} & \textbf{IoU $\uparrow$} & \textbf{MSE $\downarrow$} \\
\midrule
0  & 0.9556 & 7.45 \\
0.1  & 0.9468 & 8.75 \\
0.2  & 0.9085 & 22.87 \\
0.5  & \textit{Fail} & \textit{Fail} \\
1.0 & \textit{Fail} & \textit{Fail} \\
2.0 & \textit{Fail} & \textit{Fail} \\
\bottomrule
\end{tabular}
}

\vspace{3pt}

\end{table}

\begin{figure}[h!]
    \centering
    \includegraphics[width=\columnwidth]{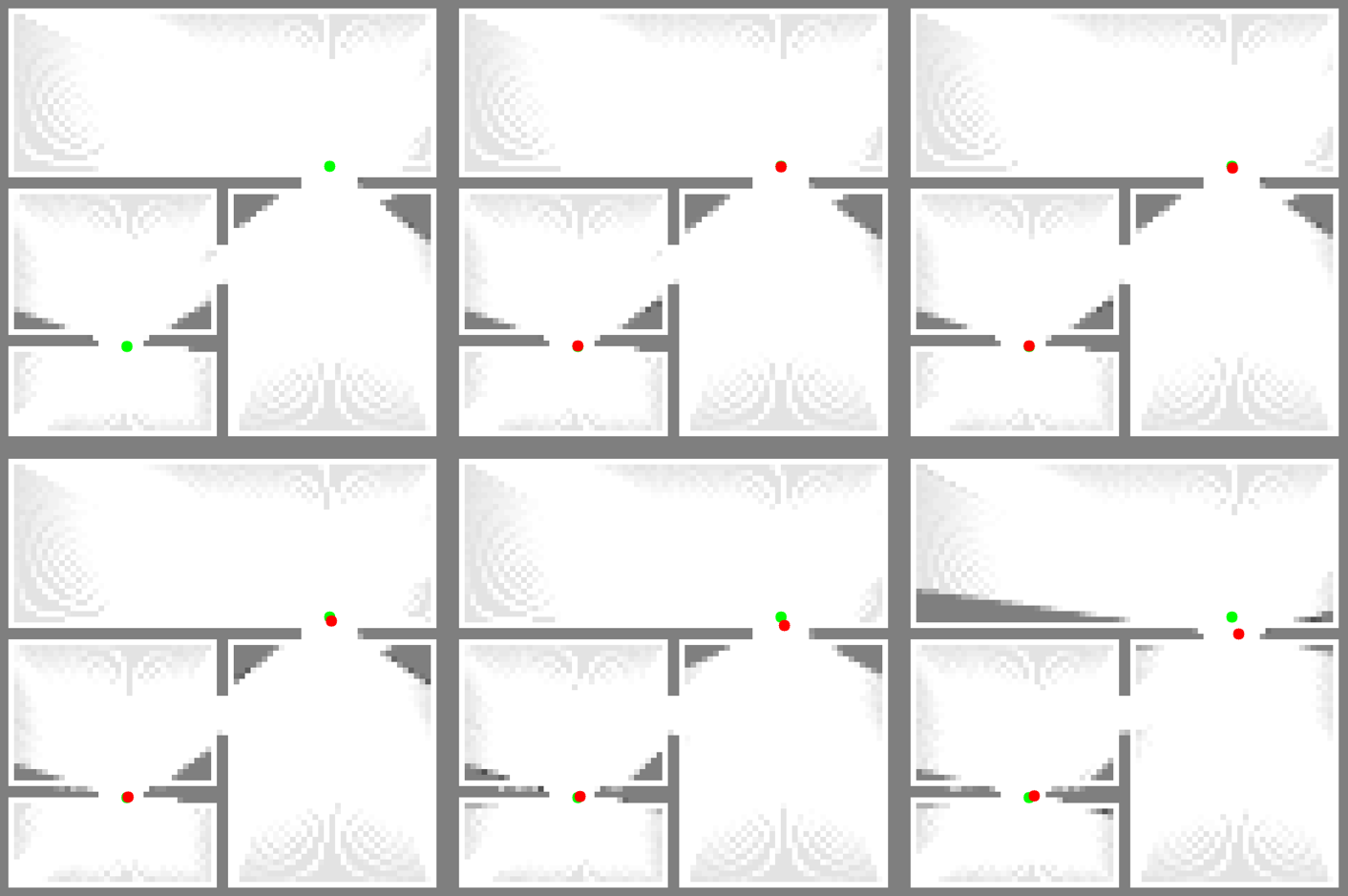}
    \caption{\textcolor{black}{Sensitivity of SfW mapping to router coordinate perturbations. Top row (left to right): $\sigma = 0$, 1, 2 px; bottom row: $\sigma = 5$, 10, 20 px. The occupancy maps remain stable with only minor degradation observed at higher noise levels. The green circles show the unperturbed positions of the routers, and the red circles show the perturbed positions of the routers.}}
    \label{fig:router_noise}
\end{figure}

\begin{figure}[h!]
    \centering
    \includegraphics[width=0.75\columnwidth]{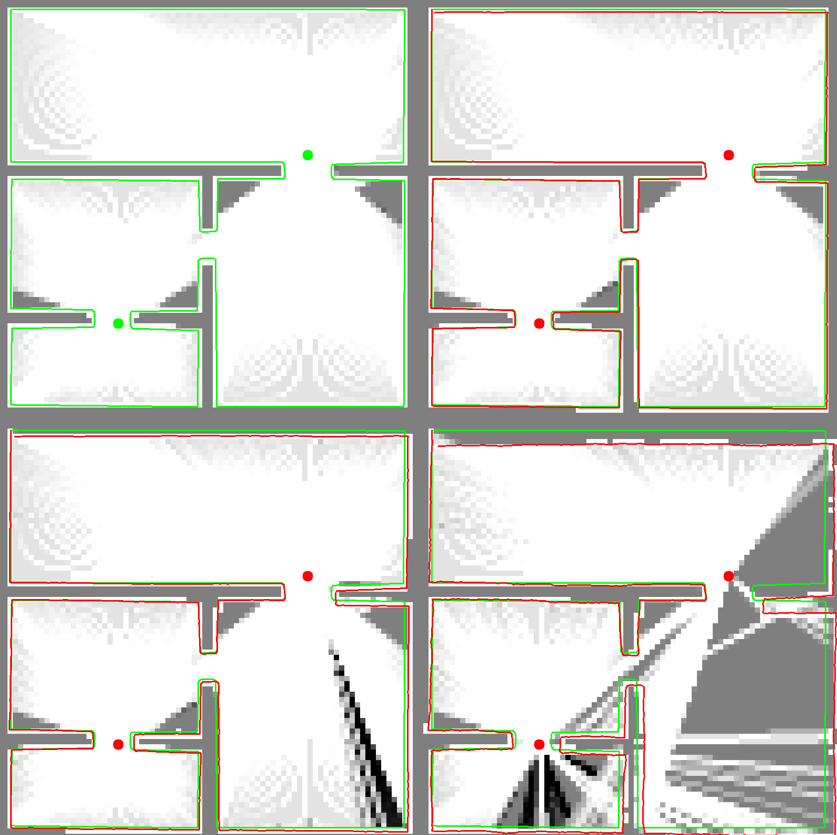}
    \caption{\textcolor{black}{Sensitivity of SfW mapping to odometry drift. Top row: $\sigma = 0$, 0.1 px; bottom row: $\sigma = 0.2$, 0.5 px. 
    The green trajectories show the unperturbed trajectory of the robot, and the red trajectories show the perturbed trajectories of the robot.
    Higher drift causes divergence and leads to failure with cases observed in Table~\ref{tab:odometry_sensitivity}.}}
    \label{fig:odom_noise}
\end{figure}

\section{Discussion and Limitations} \label{sec:dis}

\edit{In this work, we intentionally simplified the RSSI propagation model, assuming that received signal strength is primarily determined by the direct path through thick, attenuating walls, to validate the core concept of inverse k-visibility in a proof-of-concept setting. All experiments were carried out in a controlled indoor environment with thick interior walls and minimal open corridors, such that non-line-of-sight reflections, diffractions, and scattering were naturally dampened. Under these conditions, variations in RSSI could be attributed primarily to the number of intervening walls, making the direct-path approximation reasonable. This also enables the isolation and demonstration of the feasibility of RSSI-based geometric mapping without interference from multipath effects.}

\edit{However, we fully acknowledge that this modeling choice limits robustness in more general indoor scenarios. In real-world office or home environments with open-plan areas, long hallways, reflective ceilings, and dynamic agents (i.e., moving people) can dramatically alter RSSI measurements. Our current sliding window filtering is insufficient to handle these effects, and the algorithm as presented will struggle to detect and accommodate rapidly changing signal paths. Addressing these challenges will require advanced signal processing strategies such as adaptive filtering, machine learning based denoising, or incorporating additional channel metrics (i.e., multi-band analysis) to distinguish direct-path attenuation from multipath fluctuations.}

\edit{Another limitation is that our implementation is strictly 2D and planar, with both routers and trajectories confined to a ground-level grid. While this choice simplified initial validation, it does not capture ceiling-mounted access points or multi-level structures. Extending the framework to full 3D environments where vertical signal paths and signal transmission between floors become significant will be an essential next step for broader application.}

\edit{We also observed that model performance degrades once the true number of walls exceeds two (i.e., $k$ $\geq 3$). In these higher $k$ regimes, our algorithm becomes less reliable, and the algorithm's ability to accurately delineate adjacent $k$-regions diminishes. This limitation again comes from our coarse RSSI-to-$k$-mapping and the lack of explicit multipath modeling.}

\section{Future work and Conclusions} \label{sec:con}
{

\color{black}
This paper presents an innovative approach to WiFi-based geometric mapping, which effectively estimates the approximate layout of indoor spaces, especially the free space. These free spaces are necessary for the robot to plan paths and explore unfamiliar environments. Notably, our method does not depend on crowdsourcing or exteroceptive sensors such as a camera, radar, or LiDAR. Further, 
Our results demonstrate an overall increase in mapping accuracy by approximately 6.34\% on average, with about 85\% fewer data points required, compared to using a single router setup.

As for future work, it will be necessary to investigate machine learning techniques to enhance the quality of the maps during post-processing. Combining WiFi-based localization for precise trajectory tracking with our WiFi-based mapping will provide a more comprehensive autonomous WiFiSLAM system. Additionally, strategies for active trajectory planning to improve map accuracy will be explored\textcolor{black}{, including frontier-based exploration, which is navigating to frontiers between known free and unknown space, and information-gain policies, which is choosing next poses that maximize expected reduction in map uncertainty~\cite{yamauchi1997frontier, bourgault2002information, stachniss2005information, charrow2015information, bircher2016receding}}.

\textcolor{black}{More recent active SLAM methods go further by explicitly planning actions to reduce uncertainty in the SLAM belief. Approaches based on belief-space planning evaluate candidate trajectories by their expected reduction in pose/map uncertainty rather than just area coverage~\cite{vallve2015active, ahmed2023active}. In parallel, loop-closure aware planning explicitly drives the robot to re-observe past areas from new perspective to correct drift and refine the map~\cite{stachniss2004exploration}. By deliberately revisiting and closing loops during exploration, the robot can substantially lower its localization error and produce a more consistent map than purely frontier-driven strategies.}

\textcolor{black}{To eliminate residual “unknown” bands that can arise between router dominance zones (e.g., Fig.~\ref{fig:all_4_exps}-(d)), we recommend inserting short cross-connecting passes. These passes are implemented as brief perpendicular trajectories intersecting the main routes, directly traversing ambiguous or previously unseen strips between WiFi router dominance zones. By providing additional $k$=0 classification (line-of-sight) observations in these transitional areas, such passes resolve uncertain boundaries and eliminate “unknown" bands\cite{choset2001coverage,galceran2013survey}. This targeted strategy improves overall map completeness by ensuring that previously unmapped free-space segments are classified and integrated into the map.}

Further research should also focus on developing advanced \textcolor{black}{(multipath-aware} filtering systems that account for distance in the case of the large-scale map. Due to the characteristics of the RSSI signal, multiple routers should be incorporated for a large-scale environment\textcolor{black}{, preferably across multiple bands (e.g., 2.4/5 GHz) to increase robustness}. Strategic router placement is also needed for optimizing signal strength distribution and improving data quality. Routers should be positioned with minimal signal overlap and reduce areas of weak or no coverage. In most cases, this can be achieved by placing the routers at key locations such as the corners and central points of large rooms, while ensuring line-of-sight from any point.

} 


}

\section*{Acknowledgements}

We would like to thank Jill Aghyourli Zalat for her early work formulating the SfW concepts, and Ishaan Mehta for his feedback and assistance with data acquisition. This research is supported by the Natural Sciences and Engineering Research Council of Canada (NSERC).

\color{black}
\bibliographystyle{IEEEtran}
\bibliography{bibliography.bib}

\end{document}